\documentclass[11pt]{article}

\usepackage[preprint]{acl}
\usepackage[utf8]{inputenc}
\usepackage[T1]{fontenc}
\usepackage{booktabs,tabularx, multirow,xcolor}
\usepackage{ragged2e} 
\usepackage{subcaption}
\usepackage{times}
\usepackage{latexsym}
\usepackage{bm}
\usepackage{booktabs}
\usepackage{colortbl}
\usepackage{array}

\usepackage{fancyhdr}
\usepackage[T1]{fontenc}     

\usepackage{microtype}
\usepackage{tikz}
\usetikzlibrary{shapes.geometric, positioning}
\usepackage{fontawesome}
\usepackage[most]{tcolorbox}    
\usepackage{algorithm}
\usepackage{algorithmic}
\usepackage{amsmath}
\usepackage{booktabs}
\usepackage{graphicx}
\usepackage{inconsolata}
\usepackage{hyperref}
\title{Spiritual-LLM : Gita Inspired Mental Health Therapy In the Era of LLMs}

\author{%
  \textbf{Janak Kapuriya}\textsuperscript{*}\\
  Data Science Institute \\
  University of Galway, Ireland \\
  \texttt{janakkumar.kapuriya@insight-centre.org} 
  \And
  \textbf{Aman Singh}\textsuperscript{*}\\
  MIDAS Lab\\
  IIIT Delhi, India \\
  \texttt{aman22093@iiitd.ac.in} 
  \AND
  \textbf{Jainendra Shukla}\\
  HMI Lab \\
  IIIT Delhi, India \\
  \texttt{jainendra@iiitd.ac.in} 
  \And
  \textbf{Rajiv Ratn Shah}\\
  MIDAS Lab \\
  IIIT Delhi, India \\
  \texttt{rajivratn@iiitd.ac.in} 
}

\newenvironment{monopara}
  {\par\noindent}   
  {\par}

\begin{document}
\maketitle

\begingroup
  \renewcommand\thefootnote{*}%
  \footnotetext{Equal contribution.}%
\endgroup

\begin{abstract}
Traditional mental health support systems often generate responses based solely on the user's current emotion and situations, resulting in superficial interventions that fail to address deeper emotional needs. This study introduces a novel framework by integrating spiritual wisdom from the Bhagavad Gita with advanced large language models (LLM) GPT-4o to enhance emotional well-being. We present the \textbf{GITes} (Gita Integrated Therapy for Emotional Support) dataset, which enhances the existing ExTES mental health dataset by including 10,729 spiritually guided responses generated by GPT-4o and evaluated by Domain Experts.
We benchmark GITes against 12  state-of-the-art LLMs, including both mental health-specific and general-purpose models. To evaluate spiritual relevance in generated responses beyond what conventional n–gram based metrics capture, we propose a novel Spiritual Insight metric and automate assessment via an LLM-as-Jury framework using a Chain-of-Thought prompting. Integrating spiritual guidance into AI-driven support enhances both NLP and spiritual metrics for the best-performing LLM Phi3-Mini 3.2B Instruct, achieving improvements of 122.71\% in ROUGE, 126.53\% in METEOR, 8.15\% in BERT score, 15.92\% in Spiritual Insight, 18.61\% in Sufficiency and 13.22\% in Relevance compared to its zero-shot counterpart. While these results reflect substantial improvements across automated empathy and spirituality metrics, further validation in real-world patient populations remains a necessary step. Our findings indicate a strong potential for AI systems enriched with spiritual guidance to enhance user satisfaction and perceived support outcomes. The code and dataset will be publicly available to advance further research in this emerging area.

\end{abstract}

\begin{figure}[h]
    \centering
    \includegraphics[width=\linewidth]{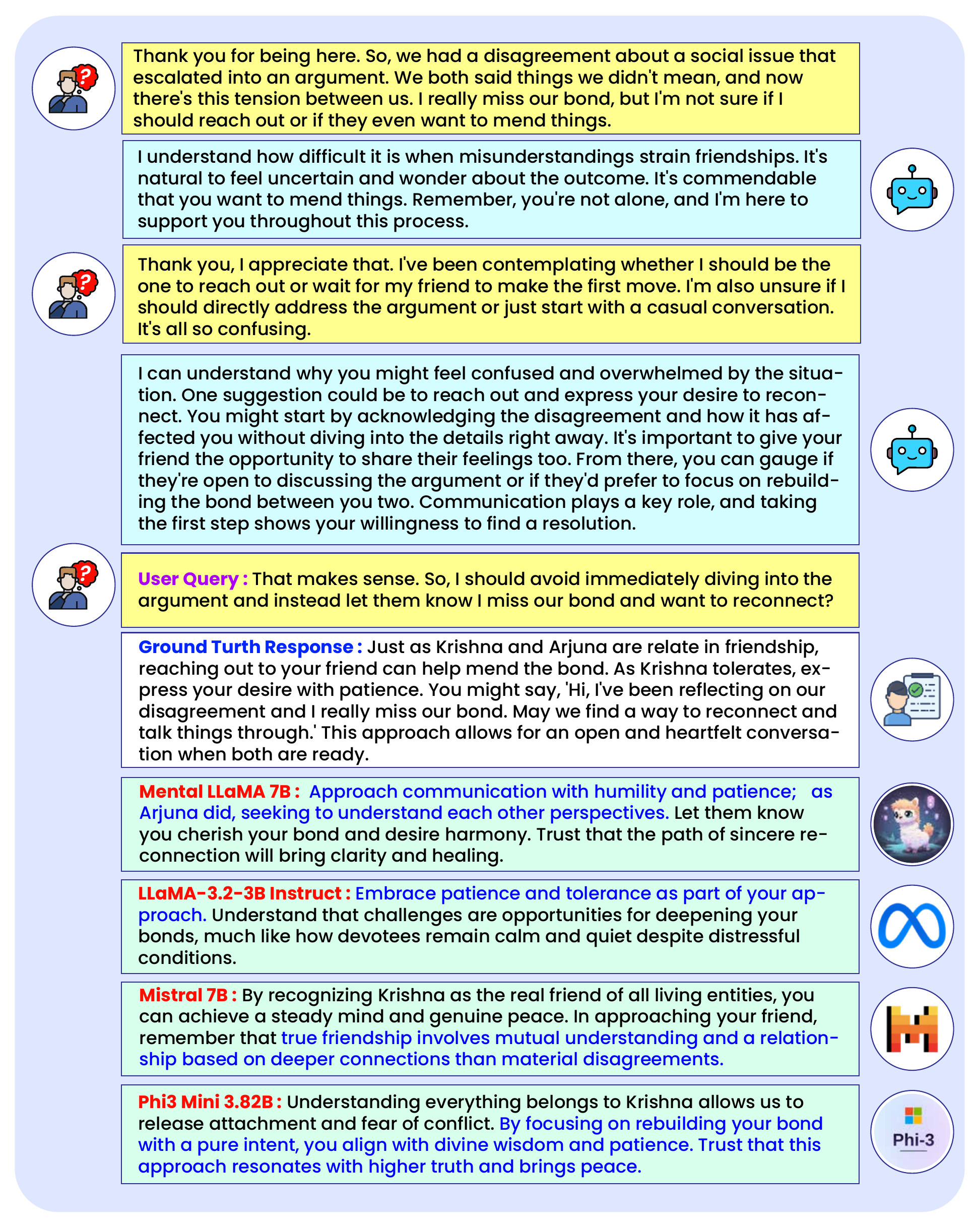}
    \caption{An example of a Context Conversation between a patient (Left) and a chatbot (right), accompanied by the patient’s query about struggling to maintain a relationship with a partner, demonstrates the capabilities of various LLMs fine-tuned on the GITes dataset. The generated responses are spiritually grounded, incorporating key teachings from the Bhagavad Gita, which are highlighted in blue.}

    \label{fig:First_Page_Example}
\end{figure}

\section{Introduction}

Mental health issues are rising worldwide, affecting all demographics and cultures. Conditions such as anxiety, depression and stress‐related disorders harm personal well‐being and carry significant social and economic costs, contributing to the global disease burden \citep{mlada2024dlouhodobe}. Modern life with its fast pace and pervasive social media—often deepens feelings of inadequacy and loneliness, creating an urgent need for accessible, empathetic support that addresses diverse emotional and spiritual needs. Furthermore, growing evidence suggests that spiritual frameworks, such as those drawn from religious texts, may provide additional dimensions of meaning-making, emotion regulation and resilience for individuals facing psychological distress. Spiritual counseling models share conceptual overlap with existential therapy, cognitive reappraisal and mindfulness-based interventions, which underscores their potential value in augmenting AI-driven mental health support.Traditional care relies on face‐to‐face sessions with psychologists and counselors, using psychotherapy, cognitive‐behavioral therapy (CBT) and pharmacotherapy to treat anxiety, depression and trauma. especially talk therapy, helps individuals express their feelings and develop coping strategies. However, challenges remain, including limited accessibility, long wait times and high costs.

The advent of artificial intelligence (AI) has revolutionized mental health care by enabling automated systems to analyze emotional states and provide support through sentiment analysis and advanced LLMs. The emergence of LLM has highlighted their remarkable ability to generalize and perform a wide range of tasks based on a limited number of examples or basic task descriptions in natural language \citep{mlada2024dlouhodobe}. Sentiment analysis utilizes natural language processing (NLP) techniques to detect and interpret emotional expressions in text, allowing for the identification of user sentiments such as happiness, sadness, anger, or anxiety \citep{abd2021perceptions}. This technology is increasingly employed in mental health applications to monitor individuals' emotional well-being in real-time, enabling timely and personalized support.

\begin{figure}[htbp]
    \centering
    \includegraphics[width=0.45\textwidth]{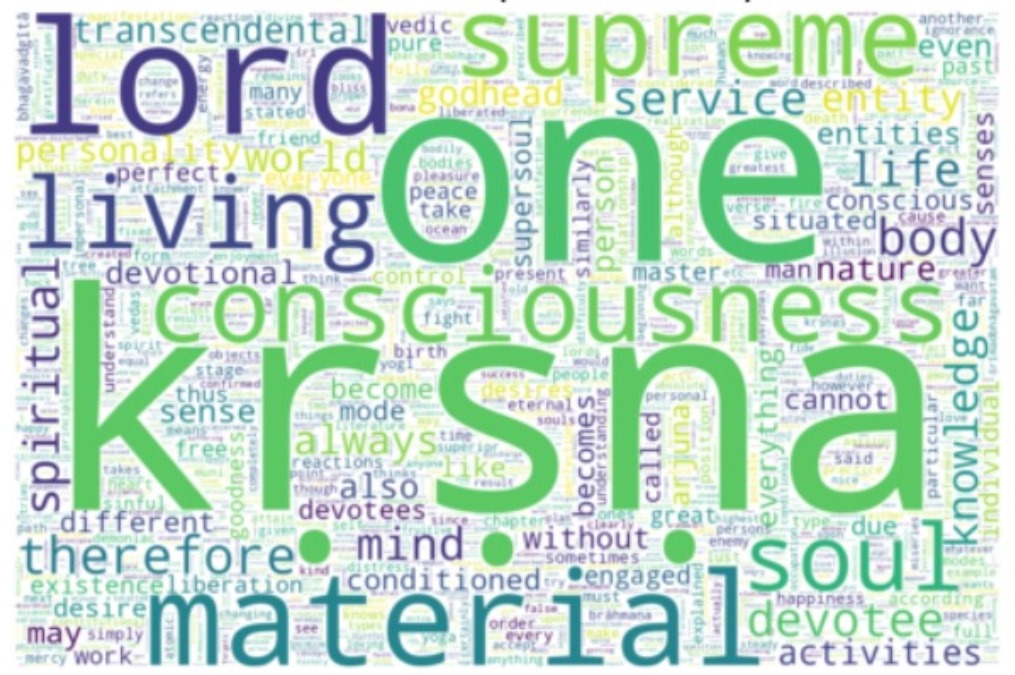}
    \caption{Word Cloud of Shloka's Purports within GITes}
    \label{fig:word_cloud_purpot}
\end{figure}

Recent LLM such as GPT4 \citep{achiam2023gpt}, GPT-4o \citep{hurst2024gpt}, DeepSeek \citep{liu2024deepseek}, Mistral \cite{ jiang2024mixtral} and  LLaMA \citep{touvron2023llama, grattafiori2024llama} 
 have shown strong instruction-following across language tasks, extending into mental health applications \citep{mann2020language,zhao2023survey}. These models can generate contextually appropriate responses based on user inputs, providing personalized emotional support and guidance. AI-driven chatbots utilizing LLM can simulate therapeutic conversations, allowing users to express their emotions and receive empathetic, constructive feedback \citep{choudhury2023investigating, fitria2023artificial}. This approach fosters connection and validation, addressing emotional needs that may not always be fully met by traditional therapies. Despite this promise, such methods risk misinterpreting emotional cues and focus solely on users’ current circumstances rather than drawing on scriptural wisdom. Incorporating guidance rooted in scripture could deliver a more holistic approach to well-being.

To bridge the gap in mental health therapy, We aim to develop and evaluate a large language model-based system that provides mental health support enhanced with spiritual insights from the Bhagavad Gita. The Bhagavad Gita, a 700-verse Hindu scripture within the Indian epic Mahabharata, features a dialogue addressing moral and philosophical dilemmas while offering spiritual guidance \citep{theodor2000philosophy}. Its teachings on emotional stability, self-realization and non-attachment align with modern therapeutic practices such as cognitive-behavioral therapy, emphasizing dharma (duty) and holistic approaches \citep{dhillon2023weaving, knott2004healing, fil2021vaishnavas}.

This study introduces GITes, a new spiritual mental health dialogue dataset that enhances the existing ExTES emotional support dataset by incorporating insights from the Bhagavad Gita into AI-generated mental health responses. ExTES covers a range of common mental health issues, such as stress, fear, anger, loneliness and confusion, etc, serves as a foundation for training LLM-based models to effectively address diverse emotional scenarios. GITes distinguishes between spiritual and non-spiritual strategies in its responses, ensuring users receive guidance that meets their emotional and spiritual needs. This GITes system can be integrated into digital mental health platforms, enhancing crisis intervention, workplace well-being programs, educational initiatives and community support with personalized spiritual guidance.

To summarize, our main contributions in this paper are as follows:

\begin{itemize}
    \item Developed the Gita‑Integrated Therapy Emotional Support \textbf{(GITes)} dataset using the \textbf{Spiritual Mental Health Dialogue Generation Framework}. By leveraging GPT‑4o, we enriched the ExTES dataset with 10,729 expert‑validated spiritual responses to support adaptable emotional dialogue systems.
    
    \item Designed a \textbf{Strategy-Based Spirituality-Aware LLM Fine-Tuning} Framework to fine-tune mental health and general LLMs on \textbf{GITes} dataset, generating spiritually-informed responses tailored to patient needs.

    \item Proposed a novel \textbf{Spiritual Insight} metric to assess spiritual context in the responses where n-gram-based methods fall short in capturing spiritual nuances. To ensure robust, reliable and automatic assessments of spiritual context, the \textbf{LLM-as-Jury Evaluation Framework} evaluates spiritual responses by aggregating ratings from three different families of Judge models.

    \item Extensive evaluations show GITes can enhance mental health counseling, social interactions and real-world applications, fostering a more compassionate society.
\end{itemize}

\section{Related Works}

\subsection{Mental Health Chatbots}
\label{sec:mental_health_chatbot_related_works}
Mental health chatbots address the global shortage of mental health professionals by offering scalable, accessible support. A review of 10 chatbots, including Woebot \citet{wan2021m} and Wysa \citet{inkster2018empathy}, analyzed user reviews (3621 on Google Play and 2624 on Apple App Store) \citet{haque2023overview}, highlighting valued human-like interactions but raising concerns about improper responses and assumptions. These chatbots primarily use CBT and mindfulness techniques, with Wysa incorporating mindfulness meditation, often seen as spiritual but not religious \citet{vaidyam2019chatbots}. Woebot offers semi-guided conversations without processing spiritual sentiments, while Wysa’s open-ended style focuses on psychological support. However, limited integration of spiritual content indicates a gap in addressing users’ spiritual needs.

\begin{figure}[htbp]
    \centering
    \includegraphics[width=0.47\textwidth] {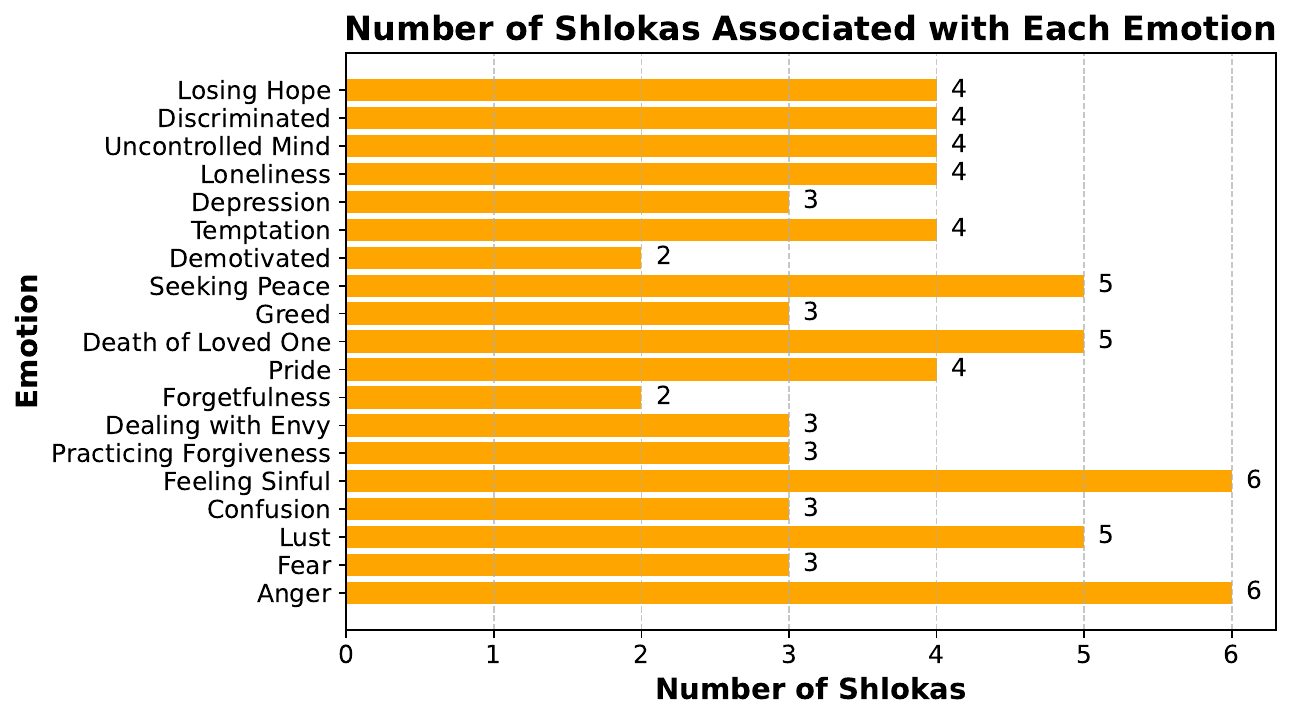}
    \caption{Emotionwise Shloka Distribution in GITes}
    \label{fig:emotions_shlokas_graph}
\end{figure}

\subsection{Emotional Dialouge Generation}

The role of emotion in constructing engaging dialogue systems has been rigorously explored in recent research 
\citep{liu2021towards}. A pivotal advancement in this domain is the Emotional Chatting Machine (ECM) proposed by \citep{liu2021towards} which generates responses aligned with predefined emotional states. It is essential to distinguish emotional chatting from Emotional Support (ES). While emotional chatting systems primarily express emotions such as happiness or sadness whereas ES systems are engineered to actively alleviate user's emotional distress through strategic conversational interventions and the deployment of advanced support methodologies
\citep{liao2023proactive, deng2023prompting, cheng2022improving}. These studies laid the groundwork for developing more sophisticated models that can understand and respond to emotional cues in various contexts, highlighting the importance of contextually rich data for enhancing emotional support systems.

\subsection{Mental Health Dialogue Datasets}
Research on emotional support conversations has involved curating and annotating datasets from diverse sources such as social media \citep{medeiros2018using}, online discussion forums \citep{sharma2020computational, hosseini2021takes} and psychotherapy session transcripts \citep{shen2020counseling}. Recent advancements demonstrate that ChatGPT outperforms traditional neural network methods in mental health analysis and model interpretability \citep{yang2024harnessing}.
To address the scarcity of mental health datasets, the Emotional Support Conversation Dataset (\textbf{ESConv}) \citep{liu2021towards} was developed using innovative data collection strategies and dialogue design. Similarly, the \textbf{ExTES} emotional support dialogue dataset \citep{zheng2023building} provides richly annotated, multi-layered scenarios emphasizing empathy and strategic intervention. The \textbf{ExTES} dataset forms the foundation of our framework. By integrating a spiritual dimension into the ExTES, we broaden its scope, enabling holistic support that bridges emotional well-being with spiritual enrichment. This integration fosters personalized guidance, creating a more effective and comprehensive emotional support system.

\subsection{Spirituality in Mental Health Care}
\label{sec:spirtuality_in_mental_health_related_works}
Research highlights the significant role of spirituality and religion in mental health care. A survey of 894 mental health professionals revealed that many consider spirituality and religion relevant to clinical practice, with clients often wanting to discuss these aspects in therapy \citet{vieten2023mental}. Incorporating spiritual and religious beliefs (SRBBPs) enhances psychological health and supports multicultural diversity. A meta-analysis of 23 randomized controlled trials found that spirituality-based therapies were moderately more effective than standard treatments for patients with strong spiritual affiliations \citet{bouwhuis2024evaluation}. Additionally, a review confirmed a consistent link between spirituality and better mental health outcomes across disorders like depression, suicidality and substance use, though mechanisms warrant further study \citet{lucchetti2021spirituality}. These findings suggest that addressing spirituality in mental health care can provide a more holistic approach, particularly for clients with strong spiritual beliefs.Integrating scriptural frameworks into therapeutic support may complement existing emotion regulation theories \citet{rottenberg2007emotion}, facilitate cognitive restructuring and align with culturally meaningful forms of acceptance and transcendence, which have demonstrated relevance across diverse mental health contexts \cite{park2013spritual}.

\subsection{LLM as Jury Evaluation}
\label{sec:llm_as_jury_related_works}
Traditional NLP metrics, such as similarity-based measures, struggle to evaluate spiritual mental health dialogues effectively. These metrics focus on surface-level alignment and fail to capture the spiritual nuance, emotional depth and contextual adaptation required for meaningful interactions. This limitation is particularly evident in mental health settings, where responses must balance empathy, relevance and emotional sensitivity. To overcome these shortcomings, recent research has explored the use of Large Language Models (LLMs) as Judge systems. For example, \citet{szymanski2024limitations} developed fine-tuning pipelines for aligning LLMs with human evaluators, while \citet{li2024llms} demonstrated their ability to assess dialogue coherence and emotional depth. Similarly, \citet{thakur2024judging} highlighted bias mitigation techniques and \citet{wang2025can} validated the approximation of human judgment in subjective tasks like mental health. These studies collectively show the potential of LLM-based evaluations for nuanced, context-aware assessments. Building on this foundation, we adopted an LLM-as-Jury Evaluation Framework for the assessment of spiritual responses by addressing the limitations of standard NLP metrics.

\begin{figure*}[h]
    \centering
    \includegraphics[width=\textwidth]{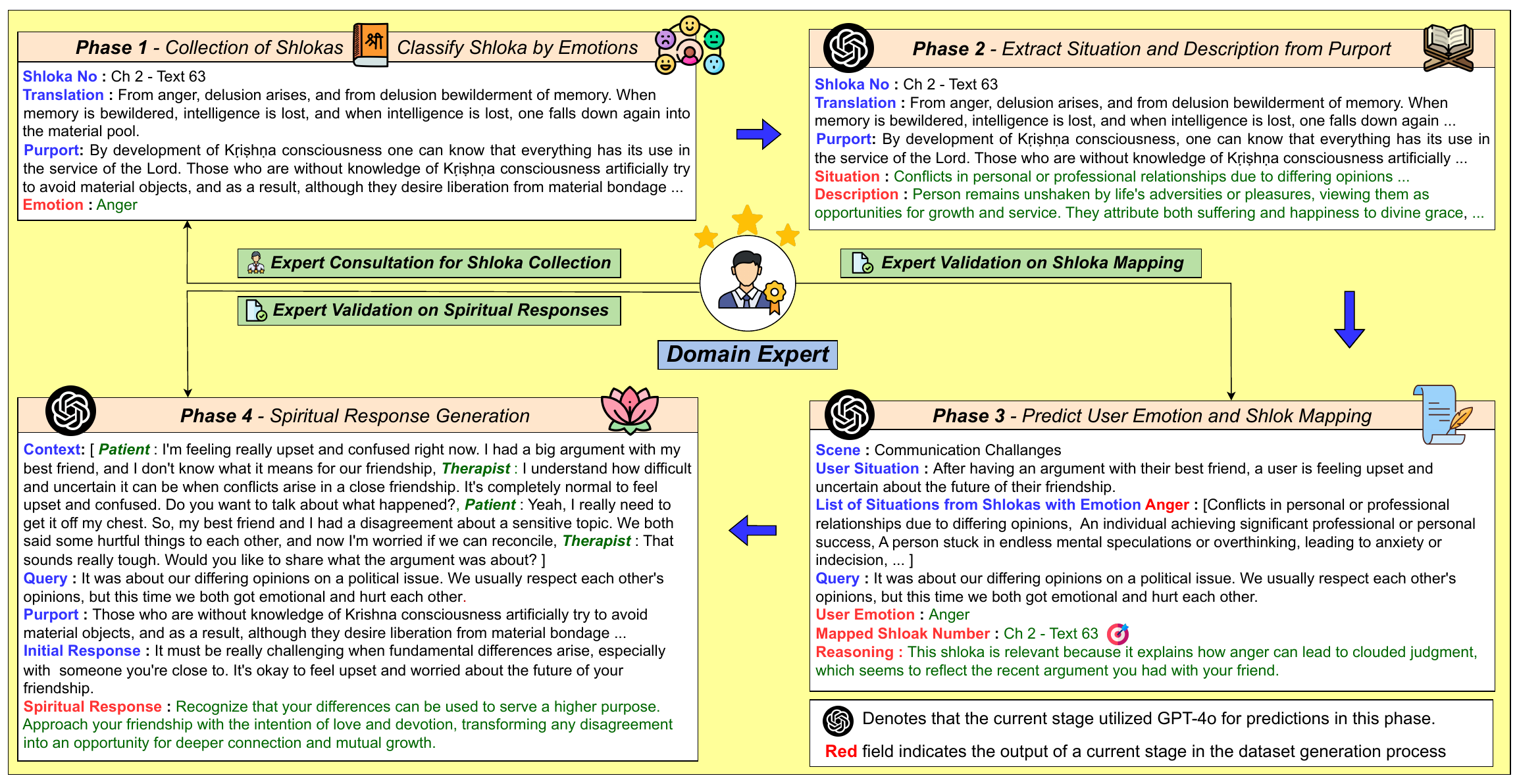}
    \caption{Schematic Representations of \textbf{Spiritual Mental Health Dialogue Generation Framework} utilized for curation of GITes Dataset}
    \label{fig:figure_3}
\end{figure*}

\section{GITes Dataset Construction} In this section, we outline our approach as shown in \textbf{Figure \ref{fig:figure_3}} to developing \textbf{GITes} an advanced, large-scale dialogue dataset for emotional and spiritual support by leveraging the generative capabilities of GPT-4o to create spiritual responses. The process is structured into four distinct phases.

\subsection{Phase 1: Collection of Shlokas}
After consulting with domain experts in Bhagawad Gita As illustrated in \textbf{Figure \ref{fig:figure_3}}, Phase 1 \citep{prabhupada1972bhagavad} from ISKCON, a globally recognized Gaudiya Vaishnava organization grounded in the teachings of the Bhagavad Gita and Srimad Bhagavatam \citep{knott2004healing}, we gathered relevant shlokas which is used by domain experts for offline counseling. These were sourced from ISKCON's authoritative publications, which are highly regarded for their deep translations and purports. Shlokas accompanied by detailed Purports, which offer contextual insights that emphasize both the spiritual and practical details for each shlokas. To enhance utility, the shlokas were categorized into specific emotions based on their relevance. Figure \ref{fig:emotions_shlokas_graph} depicts the emotion-wise distribution of the collected shlokas for GITes while Figure \ref{fig:word_cloud_purpot} shows a Word cloud of Shloka Purports.

\subsection{Phase 2: Extract Situation \& Description from Shloka Purport}
As illustrated in \textbf{Figure \ref{fig:figure_3}}, during Phase 2, we extracted the contextual Situation from the purport by leveraging a detailed prompt \footnote{\label{fn:prompt} Prompts for GITes Dataset Creation are provided in Appendix Section \ref{sec:prompts_section}} provided to GPT-4o. This prompt extracts the specific scenarios in which the purport corresponding to the shloka can be applied. Additionally, we extracted Descriptions from the purport that interpret its meaning through real-world examples. These extracted Situations form the foundation for generating meaningful and contextually relevant advice grounded in spiritual teachings.

\subsection{Phase 3: Predicting User Emotion and Shloka Mapping}
As shown in \textbf{Figure \ref{fig:figure_3}}, phase 3 comprises two key steps. First, the user's emotion is predicted based on their current query. Next, the user's Query, along with the prior conversational Context is given to GPT-4o, which identifies the most relevant shloka aligned with the user's emotional state. This is achieved by comparing the contextual conversation with the situations associated with each Purport of the shloka, which shares the same emotion as the user query. Details about the shloka mapping evaluated by the Domain Experts, along with the results, are provided in Section \ref{sec:domain_expert_evaluation}. This approach has been shown to be effective in various dataset generation tasks \citep{xu2023baize}.

\subsection{Phase 4: Spiritual Response Generation}
The framework was designed with explicit attention to clinical safety, ensuring that generated responses remain supportive, non-judgmental and avoid directive advice. Domain experts were instructed to screen for both spiritual fidelity and clinical appropriateness during evaluation stages. As illustrated in \textbf{Figure \ref{fig:figure_3}}, during Phase 4, 
the spiritual responses are generated by GPT-4o by providing a user Query, Context conversation, spiritual insights from Shloka Purpot and AI strategy with a chain-of-thought prompt. Details about the Spritual Responses evaluated by the Domain Expert, along with the results, are provided in \textbf{Section \ref{sec:domain_expert_evaluation}}. The AI strategy, as illustrated in \textbf{Figure \ref{fig:AI_Strategy_bar_graph}} determines whether to generate response spiritual or non-spiritual. Some AI strategies, such as emotional validation, reflective statements and empathetic expressions, leverage spiritual teachings to foster understanding, compassion and resilience. Conversely, other AI strategies like clarification, suggesting options and collaborative planning prioritize practical solutions without necessarily invoking spiritual insights. This distinction allows for a holistic approach to emotional support and personal development. more details about these AI strategies are given in the Appendix \textbf{Section \ref{sec:ai_strategy_ref}}.
The final \textbf{GITes} dataset comprised 10,729 spiritual conversations. This process incurred a cost of approximately \$250 for OpenAI's API usage, demonstrating a cost-efficient approach to dataset creation.

\subsection{Dataset Statistics}

Our GITes dataset has a total of 15578 samples, including both Spiritual 10729 and 4849 Non-Spiritual samples spanning across various Scene Categories, Real-World Scenarios and Diverse Emotions. During the training we use the 15578 samples and 1000 samples for the test set to evaluate model performance. More details on GITes dataset are provided in the Appendix \textbf{Section \ref{sec:statistics_appendix}}. Although the dataset was not tested in patient-facing settings, its design process incorporated iterative expert review to approximate real-world counseling dynamics while maintaining safety guardrails.

\begin{figure*}[htbp]
    \centering
    \includegraphics[width=\textwidth]{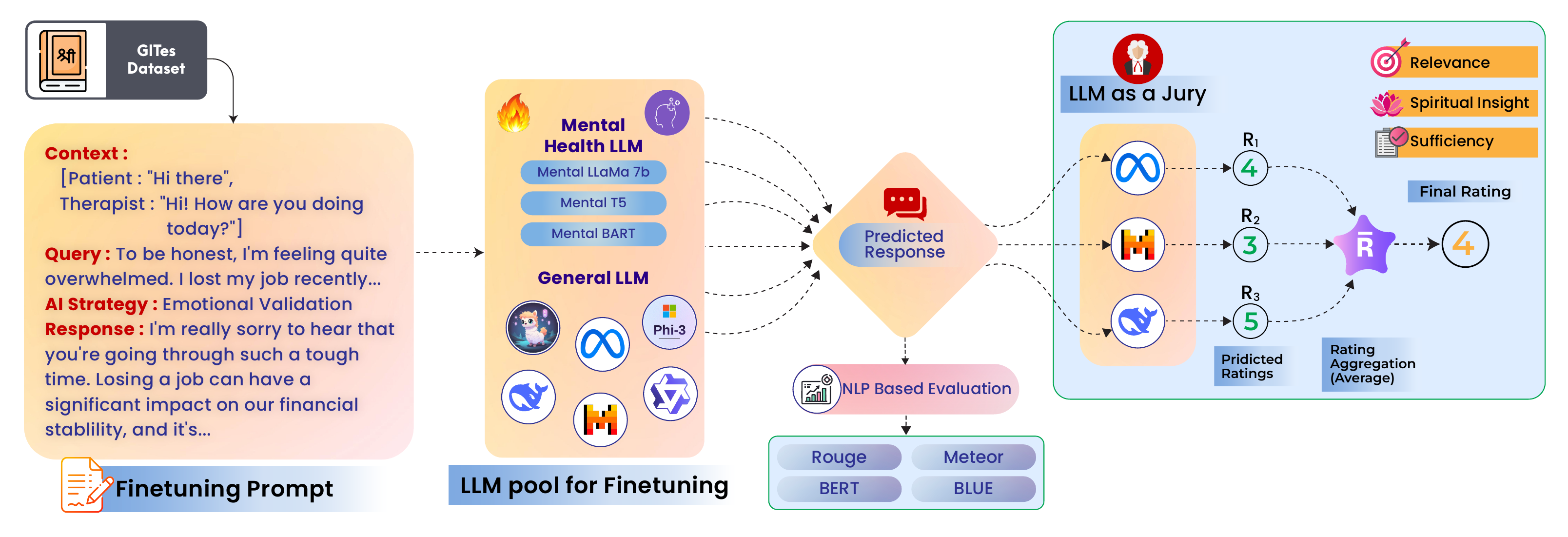}
    \caption{Strategy-Based Spirituality-Aware LLM Fine-Tuning  and LLM-as-Jury Evaluation Framework for Mental-Health Dialogue Conversations}
    \label{fig:main_diagram}
\end{figure*}

\section{Methodology}

This section outlines the Strategy-Based Spirituality-Aware LLM Fine-Tuning Framework, the evaluation metrics employed to assess response quality and the LLM-as-Jury Framework designed to automate the evaluation process detailed in the subsequent sections.

\subsection{Strategy-Based Spirituality-Aware LLM Fine-Tuning}

As Described in \textbf{Figure \ref{fig:main_diagram}}, In Finetuning Prompt, \( \mathcal{C} = (\mathcal{S}_p, \mathcal{S}_t) \) represent the Context conversation, where \( \mathcal{S}_p \) denotes the Patient's statements and \( \mathcal{S}_t \) denotes the Therapist's responses. Let \( \mathcal{Q} \) represent the Patient Query and \( \mathcal{P} \in \{0, 1\} \) represent the Preference for generating a Spiritual or Non-Spiritual response. This Preference is determined during the training based on the AI strategy as specified in Equation \ref{eq:preference}, where:

\begin{equation}
\label{eq:preference}
\mathcal{P} =
\begin{cases} 
1 & \text{for a Spiritual Response} \\
0 & \text{for a Non-Spiritual Response}
\end{cases}
\end{equation}

The sampled large language model \( \mathcal{M} \) from Pool of LLMs shown in \textbf{Figure \ref{fig:main_diagram}} generates a response \( \mathcal{R}_{\mathcal{P}} \), which could be either Spiritual (\( \mathcal{R}_s \)) or Non-Spiritual (\( \mathcal{R}_{ns} \)), as a function of the Context \( \mathcal{C} \), the Query \( \mathcal{Q} \) and the Preference \( \mathcal{P} \):

\begin{equation}
\label{eq:model_eq}
   \mathcal{R}_{\mathcal{P}} = \mathcal{M}(\mathcal{C}, \mathcal{Q}, \mathcal{P})  
\end{equation}

The training objective is to minimize the cross-entropy loss \( \mathcal{L} \) between the generated response \( \mathcal{R}_{\mathcal{P}} \) and the ground truth response \( \mathcal{R}_{\mathcal{P}}^{\text{true}} \):

\begin{equation}
\label{eq:cross_entropy}
\mathcal{L} = - \sum_{i=1}^{n} \mathcal{R}_{\mathcal{P}}^{\text{true}}(i) \log \mathcal{M}(\mathcal{C}, \mathcal{Q}, \mathcal{P})(i)
\end{equation}

Here, \( n \) represents the number of tokens in the response \( \mathcal{R}_{\mathcal{P}} \), which can either be a Spiritual (\( \mathcal{R}_s \)) or a Non-Spiritual response (\( \mathcal{R}_{ns} \)), depending on the value of \( \mathcal{P} \). Some conversations, based on the specific AI strategy, require a Spiritual response, while others demand only an empathetic response without a Spiritual element. During training, the preference for generating a Spiritual or Non-Spiritual response is assigned based on the AI strategy associated with the sample. The reasoning behind selecting a particular AI strategy is detailed in Appendix \textbf{Section \ref{sec:ai_strategy_ref}}. We use a detailed prompt given in \textbf{Figure \ref{fig:llm_finetuning_prompt}} to generate contextually rich and spiritually relevant responses. During fine-tuning \textbf{(Equation \ref{eq:model_eq})} Cross-Entropy Loss \textbf{(Equation \ref{eq:cross_entropy})} minimizes prediction errors, ensuring coherent, accurate and spiritually enriched dialogues, fostering meaningful interactions in mental health settings.


\section{Evaluation Framework}
\label{subsec:eval_metrics}

\subsection{Standard NLP Metrics} We evaluated the generated mental health responses using standard NLP metrics, including \textbf{ROUGE} \citep{lin2004rouge}, \textbf{BERTScore} \citep{zhang2019bertscore}, \textbf{BLEU} \citep{papineni-etal-2002-bleu} and \textbf{METEOR} \citep{banerjee2005meteor}. ROUGE score measures key phrase overlapping, BERTScore captures semantic similarity, while BLEU score evaluates precision of the generated text and METEOR assesses word alignment with synonym to evaluate predicted responses with ground truth. These metrics collectively provide an assessment of response quality and relevance. To further strengthen comparative rigor, we performed non-parametric bootstrap resampling (N=1000) to derive confidence intervals for each metric. Pairwise comparisons were conducted between fine-tuned models and their zero-shot counterparts to quantify the robustness of observed improvements.

\subsection{Spiritual Metrics} We proposed one novel evaluation metric \textbf{Spiritual Insight}, while utilized two existing general metrics \textbf{Sufficiency} and \textbf{Relevance} in spiritual context to assess the generated responses. where \textbf{Spiritual Insights} evaluates the depth and accuracy of the spiritual guidance in the response, based on principles derived from the Gita shlokas. \textbf{Sufficiency} evaluates the completeness of the response in addressing the patient’s query within the conversation context. \textbf{Relevance} measures how well the response addresses the patient's query within the context of the conversation. We initially used Cosine similarity over embeddings yielded uniformly high scores across distinct responses, failing to capture relevance, sufficiency, or spiritual insight due to embedding saturation, geometric biases and lack of contextual nuance \cite{reimers2019sentence,ethayarajh2019contextual,zhou2022problems}. further show that it can systematically misrepresent similarity for frequent term \textbf{(Figure \ref{fig:word_cloud_purpot})} due to embedding space distortion. To overcome these limitations, we adopted the LLM-as-Jury Framework with a Chain-of-Thought prompt  evaluation, enabling fine-grained, interpretable judgments better aligned with human assessments \cite{liu2023pre, szymanski2024limitations}.

\subsection{Domain Expert Evaluation}
\label{sec:domain_expert_evaluation}

The domain expert evaluation was conducted during the creation of the GITes dataset to assess the accuracy of Shloka Mapping and Ground Truth Spiritual Responses generated by our spiritual mental health response generation framework. The shloka mapping was evaluated by three domain experts using a random sample of 100 instances on the rating scale of (1-3) categorized as Low, Medium and High Mapping. Ground Truth spiritual response evaluated by three domain experts on randomly sampled 50 instances on the rating scale of (1-3) categories as Low, Medium and High. The Expert Agreement Score, measured using Cohen's Kappa, is presented in \textbf{Table \ref{tab:cohen_kappa_table}}. The scores range from 0.2 to 0.4, indicating a fair level of agreement among experts. These results suggest that the responses generated by GPT-4o with mapped shlokas and their teaching are effectively aligned with user queries, providing accurate and contextually relevant guidance.

\begin{table}[h]
\centering
\renewcommand{\arraystretch}{1.3}
\begin{tabular}{lcc}
\toprule
\textbf{Parameter} & \textbf{\bm{$\kappa$}} & \textbf{Agreement Level} \\
\midrule
Shlok Mapping    & 0.245 & Fair \\
Spiritual Responses & 0.261 & Fair \\
\bottomrule
\end{tabular}
\caption{Cohen's Kappa Score ($\kappa$) indicating Annotators' Agreement}
\label{tab:cohen_kappa_table}
\end{table}


\subsection{LLM-as-Jury Evaluation Framework}
To evaluate Spiritual responses on domain-specific Spiritual Metrics such as Spiritual Insight, Sufficiency and Relevance, as explained in \textbf{Section \ref{sec:llm_as_jury_related_works}}, we employ the LLM as Jury framework.

We denote the predicted response for a test sample $i \in \{1, 2, \dots, N\}$ as $\mathcal{R}_{\mathcal{P}_i}$, where $N$ is the total number of samples in the test set. The evaluation process involves three judge models: \textbf{LLaMA}, \textbf{Mistral} and \textbf{DeepSeek}, following these steps as shown Right side of the \textbf{Figure \ref{fig:main_diagram}}:

\begin{enumerate}
    \item Each predicted test response $\mathcal{R}_{\mathcal{P}_i}$ is evaluated by the judge models using a Chain of Thought Prompting\footnote{Prompt is provided in Appendix for Spiritual Insight in Fig \textbf{\ref{fig:spiritual_insight_jury_prompt}}, Relevance in Fig \textbf{\ref{fig:relevence_jury_prompt}} and \textbf{Sufficiency in Fig \ref{fig:sufficiency_prompt}}} approach, generating ratings:
    \begin{align*}
        R_{1,i} &: \text{LLaMA 3.1 8B Instruct}, \\
        R_{2,i} &: \text{Mistral 7B Instruct}, \\
        R_{3,i} &: \text{Deepseek R1 Distill LLaMA 8B}.
    \end{align*}
    \item The aggregated rating $\bar{R}_i$ for sample $i$ is computed as:
    \begin{equation}
        \bar{R}_i = \frac{R_{1,i} + R_{2,i} + R_{3,i}}{3}.
    \end{equation}
    \item This process is repeated for all $N$ samples, resulting in aggregated ratings $\{\bar{R}_1, \bar{R}_2, \dots, \bar{R}_N\}$.
\end{enumerate}

These evaluation steps were conducted for three Spiritual Metrics: \textbf{Spiritual Insight}, \textbf{Sufficiency} and \textbf{Relevance}. The overall performance for each metric $M_k$ is represented by the mean aggregated rating $\bar{R}_{M_k}$:
\begin{equation}
    \bar{R}_{M_k} = \frac{1}{N} \sum_{i=1}^N \bar{R}_{i,M_k}.
\end{equation}

This mean rating provides a comprehensive assessment of the model's performance over the test set under the "LLM as Jury" framework.

\begin{table*}[h]
\centering
\resizebox{0.95\linewidth}{!}{
\begin{tabular}{lccccccc}
\toprule
\textbf{Model} & \multicolumn{4}{c}{\textbf{Standard NLP Metrics}} & \multicolumn{3}{c}{\textbf{Spiritual Metrics}} \\
\cmidrule(lr){2-5} \cmidrule(lr){6-8}
 & \textbf{ROUGE-L} & \textbf{BLEU-4} & \textbf{METEOR} & \textbf{BERT} & \textbf{Spiritual Insight} & \textbf{Sufficiency} & \textbf{Relevance} \\
\midrule
\rowcolor{gray!20} \multicolumn{8}{c}{\textbf{Mental Health Language Models}} \\

Mental BART & 14.97 \textsuperscript{$\alpha,\beta$} & 1.05 \textsuperscript{$\alpha,\beta$} & 14.30 \textsuperscript{$\alpha$}& 84.77 \textsuperscript{$\alpha,\beta$}  & 2.31 \textsuperscript{$\alpha,\beta$} & 2.17 \textsuperscript{$\alpha,\beta$}  & 2.19 \textsuperscript{$\alpha,\beta$}\\
Mental LLaMA 7B Chat & 20.05\textsuperscript{$\alpha,\beta$} & 3.30\textsuperscript{$\alpha,\beta$} & 22.21\textsuperscript{$\alpha,\beta$} & 87.18\textsuperscript{$\alpha,\beta$}  & \textbf{3.67}\textsuperscript{$\alpha,\beta$} & \textbf{3.36}\textsuperscript{$\alpha,\beta$} & \textbf{3.14} \textsuperscript{$\alpha,\beta$}  \\
Mental T5 & \textbf{24.64}\textsuperscript{$\alpha,\beta$} & \textbf{5.02}\textsuperscript{$\alpha,\beta$} & \textbf{24.61}\textsuperscript{$\alpha,\beta$} & \textbf{87.38}\textsuperscript{$\alpha,\beta$}  & 3.24 \textsuperscript{$\alpha,\beta$} & 2.85 \textsuperscript{$\alpha,\beta$} & 2.79 \textsuperscript{$\alpha,\beta$} \\
\midrule
\rowcolor{gray!20} \multicolumn{8}{c}{\textbf{General Large Language Models}} \\
DeepSeek-R1-Distill-Qwen-7B & 17.13 \textsuperscript{$\alpha,\beta$} & 2.18 \textsuperscript{$\alpha,\beta$} & 18.72 \textsuperscript{$\alpha,\beta$} & 84.26 \textsuperscript{$\alpha,\beta$} & 3.66 \textsuperscript{$\alpha,\beta$}  & 3.54 \textsuperscript{$\alpha,\beta$} & 3.03  \textsuperscript{$\alpha,\beta$} \\
Qwen2 7B Instruct & 17.40 & 2.53  \textsuperscript{$\alpha,\beta$} & 18.83\textsuperscript{$\alpha,\beta$} & 85.77\textsuperscript{$\alpha,\beta$}  & 3.69\textsuperscript{$\alpha,\beta$} & 3.32\textsuperscript{$\alpha,\beta$} & 2.96\textsuperscript{$\alpha,\beta$}\\
LLaMA 3.1 8B Instruct & 18.15 \textsuperscript{$\alpha,\beta$} & 2.81 \textsuperscript{$\alpha,\beta$} & 19.50  & 85.87 \textsuperscript{$\alpha,\beta$} & 3.49 \textsuperscript{$\alpha,\beta$} & 3.22  & 2.98 \textsuperscript{$\alpha,\beta$} \\
DeepSeek-R1-Distill-LLaMA-8B & 19.36 \textsuperscript{$\alpha,\beta$} & 3.02 \textsuperscript{$\alpha,\beta$} & 20.91 \textsuperscript{$\alpha,\beta$} & 84.74 \textsuperscript{$\alpha,\beta$} & 3.81 \textsuperscript{$\alpha,\beta$} & 3.60 \textsuperscript{$\alpha$} & 3.02 \textsuperscript{$\alpha,\beta$}\\
LLaMA 2 7B Chat & 19.98\textsuperscript{$\alpha,\beta$} & 3.46\textsuperscript{$\alpha,\beta$} & 22.05\textsuperscript{$\alpha,\beta$} & 87.15\textsuperscript{$\alpha,\beta$}  & 3.64\textsuperscript{$\alpha,\beta$} & 3.34\textsuperscript{$\alpha,\beta$} & 3.07\textsuperscript{$\alpha,\beta$} \\
Mistral 7B Instruct & 20.76\textsuperscript{$\alpha,\beta$} & 3.73\textsuperscript{$\alpha,\beta$} & 23.05 \textsuperscript{$\alpha,\beta$} & 87.39 & 3.74 \textsuperscript{$\alpha,\beta$} & 3.42 \textsuperscript{$\alpha,\beta$} & 3.19 \textsuperscript{$\alpha,\beta$}\\
DeepSeek-R1-Distill-Qwen-1.5B & 26.48\textsuperscript{$\alpha,\beta$} & 9.01\textsuperscript{$\alpha,\beta$} & 28.15\textsuperscript{$\alpha,\beta$} & 85.76\textsuperscript{$\alpha,\beta$}  & 3.88\textsuperscript{$\alpha,\beta$} & 3.71\textsuperscript{$\alpha,\beta$} & 3.11\textsuperscript{$\alpha,\beta$} \\
Phi 3.5 Mini 3.82B Instruct
 & 27.61 \textsuperscript{$\alpha,\beta$} & 10.11 \textsuperscript{$\alpha,\beta$} & \textbf{29.87}\textsuperscript{$\alpha,\beta$} & \textbf{88.93} \textsuperscript{$\alpha,\beta$} & \textbf{4.03}\textsuperscript{$\alpha,\beta$} & \textbf{3.76}\textsuperscript{$\alpha,\beta$} & \textbf{3.64}\textsuperscript{$\alpha,\beta$} \\
LLaMA 3.2 3B Instruct & \textbf{28.19} \textsuperscript{$\alpha,\beta$} & \textbf{10.22} \textsuperscript{$\alpha,\beta$} & 29.85 \textsuperscript{$\alpha,\beta$} & 87.42  \textsuperscript{$\alpha,\beta$} & 3.63 \textsuperscript{$\alpha,\beta$} & 3.38 \textsuperscript{$\alpha,\beta$} & 3.10 \textsuperscript{$\alpha,\beta$} \\
\hline
\bottomrule
\end{tabular}
}
\caption{Results for various LLMs fine‑tuned on the GITes dataset, evaluated using standard NLP and spiritual metrics. Superscript $\alpha$ denotes statistical significance on pairwise permutation test on ZS vs SFT at the 5\% confidence Interval (CI), while $\beta$ represents at the 1\% confidence Interval (CI).}
\label{tab:finetuning_results_table}
\end{table*}


\section{Experiments}
\label{sec:experiments}
To address mental health challenges through spiritual integration, we utilize specialized mental health language models alongside general-purpose large language models (LLMs) with proven zero-shot capabilities. Our experimental framework combines comparative analysis of domain-specific and general models under different training paradigms.

\subsection{Mental Health Specialized Models} We selected three domain-specific language models Mental-BART, Mental-T5 and Mental-LLaMA-7B-Chat \citep{yang2023mentalllama}, pretrained on mental health data. These models, optimized through pretraining on clinical dialogues and therapeutic datasets, were benchmarked on GITes to evaluate their performance in mental health dialogue generation.

\subsection{General LLM} To assess adaptability for spiritually-aware mental health conversations, we benchmark 9 recent open-source LLMs across five different family of models. These include Mistral-7B-Instruct \citep{jiang2023mistral}, three LLaMA variants (LLaMA2-7B-Chat, LLaMA3.2-3B-Instruct and LLaMA3.1-8B-Instruct) \citep{touvron2023llama, dubey2024llama}, three DeepSeek variants (DeepSeek-R1-Distill-Qwen-1.5B, DeepSeek-R1-Distill-Qwen-7B and DeepSeek-R1-Distill-LLaMA-8B) \citep{liu2024deepseek}, Qwen2-7B-Instruct \citep{yang2024qwen2} and Phi3.5-Mini-Instruct \citep{abdin2024phi}.

\subsection{Experimental Setup} We conducted two experiments \textbf{1) Zero-Shot Prompting} and \textbf{2) Supervised Fine-Tuning} to evaluate the efficacy of the GITes dataset in integrating spiritual elements into mental health dialogues. In the zero-shot setting, we directly prompted each model to generate responses, assessing its inherent spiritual ability to address patient queries. In the fine-tuning phase, using the GITes dataset, we adapted the models to incorporate spiritual dimensions into therapeutic conversations. To optimize resource utilization, we employed the Low-Rank Adaptation (LoRA) technique \citep{qin2024accurate} across all models, except for Mental T5 and Mental BART, during both experimental phases. This ensured efficient memory management given our GPU constraints. more details about hyperparameters used during this experiments were given in the Appendix \textbf{Section \ref{sec:hyperparameters}}.

\section{Results}
\label{ref:results}

\subsection{Main Results}
We present the fine-tuning results in \textbf{Table \ref{tab:finetuning_results_table}}, evaluated using both standard NLP and spiritual metrics, for mental health-specific models and general-purpose LLMs. Among the mental health-specific models, Mental-T5 achieves the highest scores in standard NLP metrics, with 24.645\% Rouge-L, 5.029\% BLEU-4, 24.613\% METEOR and 87.387\% BERT score, highlighting its strong linguistic quality in response generation. On the other hand, Mental-LLaMA excels in spiritual metrics, achieving Rating score of 3.67 in Spiritual Insight, 3.36 in Sufficiency and 3.14 in Relevance, demonstrating its superior domain-specific adaptation for providing spiritual guidance. These results illustrate a balance between linguistic precision and spiritual relevance. As we can see in \textbf{Figure \ref{fig:error_example_3},} Mental-LLaMA provides not just information, but a spiritual framing consistent with the tone of the Gita on long long-distance relationship query of patient.\\

Compact general-purpose LLMs, such as Phi3.5 Mini 3.8B and LLaMA-3.2 3B, perform exceptionally well on spiritual metrics, showcasing that smaller architectures with strong instruction-following capabilities can thrive in specialized domains while minimizing computational costs. Among these, Phi3.5 Mini stands out by achieving the highest scores across all spiritual metrics, with ratings of 4.03 in Spiritual Insight, 3.76 in Sufficiency and 3.64 in Relevance. Additionally, it leads in two NLP metrics, with a METEOR score of 29.872\% and a BERT score of 55.938\%, while securing the second-best results in ROUGE-L and BLEU-4. The top performer for ROUGE-L and BLEU-4 is LLaMA-3.2 3B, with scores of 28.198\% and 10.220\%, respectively. These findings highlight Phi3.5 Mini as an ideal model for generating spiritually nuanced and contextually relevant responses.  As shown in \textbf{Table  \ref{tab:good_example_table}}, both Phi-3 and LLaMA-3.2 models provide a response grounded in Gita's teachings to address the patient's query about a long-distance relationship, with key teachings highlighted in blue.

\begin{figure}[htbp]
    \centering
    \includegraphics[width=0.48\textwidth]{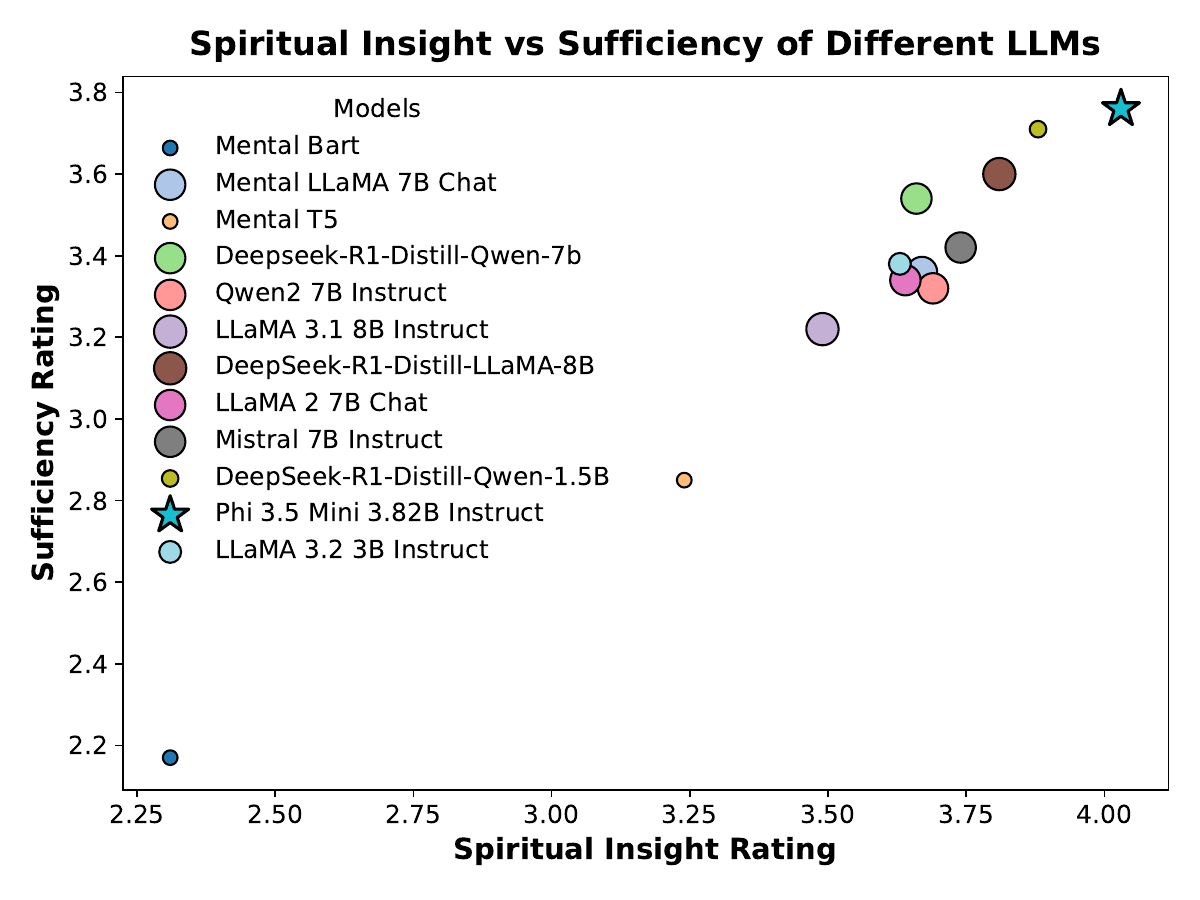}
    \caption{Comparison of \textbf{Spiritual Insight} vs \textbf{Sufficiency} performance of different LLMs on LLM-as-Jury Evaluation Framework}
    \label{fig:Spiritual_Insight_vs_Sufficiency_plot}
\end{figure}

Larger models, such as DeepSeek-R1-Distill-Qwen-7B and Qwen2-7B, perform lower on NLP metrics compared to the best model, Phi3.5 Mini. However, they achieve competitive scores in Spiritual Insight and Sufficiency, highlighting the challenges of spiritually and emotionally sensitive domains, where nuanced understanding and insight are critical. \textbf{Figure \ref{fig:Spiritual_Insight_vs_Sufficiency_plot}} presents a scatter plot illustrating the strong positive correlation between Spiritual Insight and Sufficiency. Among the models evaluated, Phi-3 Mini stands out as the best-performing model on the spiritual metric, while Metal Bart demonstrates the lowest performance. This highlights the superior instruction-following capabilities of advanced models. Bootstrap analysis confirmed that the observed improvements across both NLP and spiritual metrics were statistically robust (95\% CI), with the Phi3-Mini model achieving consistent superiority over baseline configurations. Full pairwise permutation test Confidence Interval scores available in Supplementary Appendix \textbf{Table \ref{tab:spiritual_metric_Permutation_test_scores}} and \textbf{Table \ref{tab:NLP_Metric_Permutation_test_scores}}.

\begin{figure*}[htbp]
    \centering
    \includegraphics[width=\textwidth]{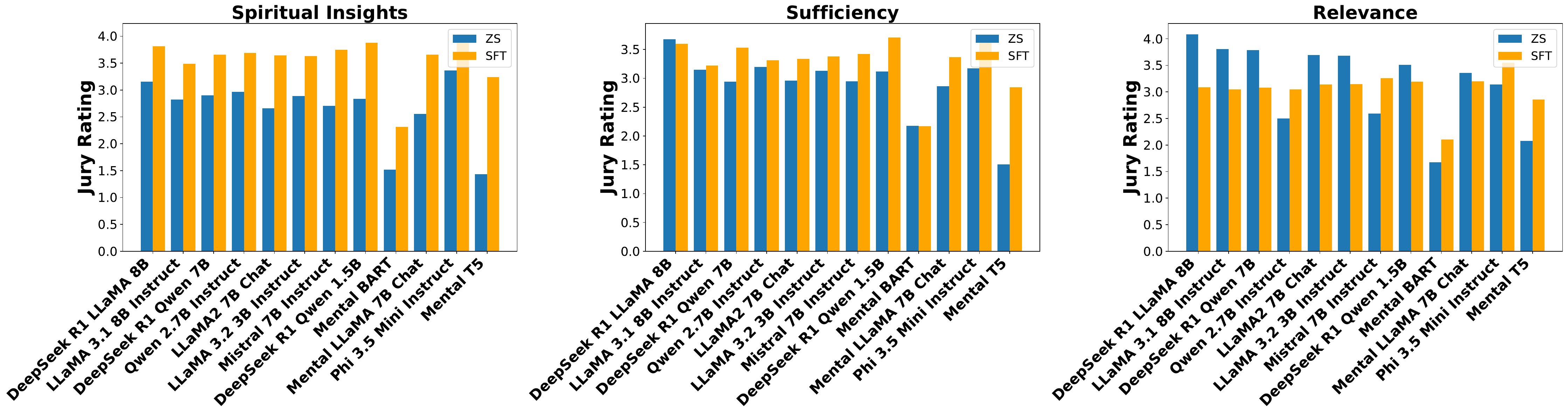}
    \caption{Comparison of \textbf{Zero Shot} vs \textbf{Supervised Finetuning} Performance of different LLMs on Spiritual Metrics with LLM-as-Jury Evaluation Framework.}
    \label{fig:ZS_vs_SFT_bar_graph}
\end{figure*}

\subsection{Zero-Shot vs SFT Analysis} 
We compared Zero-Shot (ZS) prompting and Supervised Fine-Tuning (SFT) across three spiritual metrics, as shown in \textbf{Figure \ref{fig:ZS_vs_SFT_bar_graph}}. On Spiritual Insight, all models performed better with SFT, underscoring the critical role of the GITes dataset in providing meaningful spiritual context. Mental health-specific models exhibited the largest performance boost, demonstrating their strong domain adaptation to spiritual contexts. For Sufficiency, most models showed significant improvement with fine-tuning. Notably, all mental health models except Mental BART and all general LLMs except DeepSeek-R1-LLaMA-8B, outperformed their Zero-Shot counterparts. These findings suggest that fine-tuned models are better equipped to generate comprehensive responses. In contrast, the impact of SFT on Relevance was less consistent, with only 5 out of 13 models improving over ZS performance. These included Mental T5, Mental BART, Qwen-2.7 7B, Mistral 7B and Phi3.5 Mini, highlighting the strengths of domain-specific models and the robust instruction-following capabilities of certain general-purpose LLMs. Further results for individual judge models on ZS vs. SFT are detailed in \textbf{Figure \ref{fig:ZS_vs_SFT_results_Judge_models}}. These findings suggest that fine-tuning not only optimizes linguistic fluency but also facilitates more nuanced integration of scriptural teachings, potentially enhancing users' sense of spiritual resonance and meaning-making mechanisms linked to therapeutic alliance and positive emotional reappraisal in prior psychotherapy literature\cite{vieten2013spiritual}. Overall, while SFT consistently enhances Spiritual Insight and Sufficiency, additional refinement is needed to improve Relevance, particularly for addressing specific patient needs in mental health contexts. To gain deeper insights into these quantitative findings, we next examine real-world dialogue examples and failure modes in detail.

\begin{table}[h]
\centering
\large
\resizebox{\columnwidth}{!}{%
\begin{tabular}{lccc}
\toprule
\textbf{Model} & \textbf{Spiritual Insight} & \textbf{Sufficiency} & \textbf{Relevance} \\
\midrule
Mental LLaMA 7B Chat & 3.660 & 3.365 & 3.199 \\
LLaMA3-3.2B Chat & 3.633 & 3.383 & 3.149 \\
Phi3-Mini 2.7B Chat & \textbf{4.035} & \textbf{3.763} &  \textbf{3.637} \\
\hline
Average & 3.776 & 3.503 & 3.328  \\
\bottomrule
\end{tabular}%
}
\caption{Human-as-Jury Evaluation Results on Three Spiritual Metrics Across Three Different LLMs, Including Mental Health and General Purpose Models.}
\label{tab:human_as_jury_eval_results}
\end{table}

\subsection{Human-as-Jury Evaluation}
To evaluate the alignment between the LLM-as-Jury Framework and Human Judgment, we conducted an experiment involving three domain experts. The experts rated 10 randomly selected samples across three spiritual metrics. The average scores from this evaluation on three different LLMs are presented in \textbf{Table \ref{tab:human_as_jury_eval_results}}.
The results reveal a consistent trend with the LLM-as-Jury evaluations, as depicted in \textbf{Figure \ref{fig:ZS_vs_SFT_bar_graph}}. Among the metrics, Spiritual Insight received the highest ratings, followed by Sufficiency, with Relevance receiving the lowest scores. This alignment demonstrates that the LLM-as-Jury Framework provides a reliable approach to automate large-scale evaluation.

\definecolor{HeaderBG}{named}{yellow}

\definecolor{CompBG}{gray}{0.82}

\newcolumntype{M}[1]{>{\Centering\arraybackslash}m{#1}}
\newcolumntype{L}[1]{>{\RaggedRight\arraybackslash}m{#1}}

\setlength{\extrarowheight}{4pt}

\begin{table*}[t]
  \centering
  \small
  \begin{tabular}{|
      >{\columncolor{CompBG}}L{0.20\linewidth}
      |L{0.75\linewidth}|
    }
    \hline
    \rowcolor{HeaderBG}
    \textbf{Component} & \textbf{Good Example} \\
    \hline
    \textbf{Context Conversation} &
      \textbf{\textcolor{purple}{Patient:}} I guess, I just needed someone to talk to and share my grief. I feel guilty for not realizing he was sick sooner. \\
    & \textbf{\textcolor{purple}{Therapist:}} It's very common to feel guilty in these situations, but please remember that you provided a loving home for him. Dogs are very good at hiding their pain, so it's natural to not notice any signs. Blaming yourself won't bring him back, but acknowledging the love and care you gave him can bring some solace. \\
    & \textbf{\textcolor{purple}{Patient:}} You're right. I just wish there was something more I could have done. \\
    & \textbf{\textcolor{purple}{Therapist:}} Losing a pet is never easy and it's normal to have those thoughts. But try to focus on the happy memories you shared with him. Celebrate his life and the happiness he brought you. And know that he felt your love throughout his time with you. \\
    \hline
    \textbf{Patient Query} & \textbf{I will try to do that. Thank you for reminding me of the good times.} \\
    \hline
    \textbf{Ground Truth} &
      \textbf{\textcolor{orange}{Practice patience and be kind to yourself during this grieving process. Embrace the idea that difficulties can be a form of the Lord's mercy, guiding you closer to spiritual understanding and peace.}} \\
    \hline
    \textbf{Mental LLaMA 7B} &
      Remember, \textbf{\textcolor{blue}{true peace comes from recognizing Lord Krishna as the supreme benefactor. Trust that He offers both nearness and farness to us}}; acknowledge His role in your life and celebrate the bond you shared. \\
    \hline
    \textbf{LLaMA-3.2-3B Instruct} &
      Remember that \textbf{\textcolor{blue}{every living being is an individual soul and like the atomic soul, the spirit soul does not undergo any change.}} Cherish the love and joy he brought you. Finding peace in knowing this eternal truth helps us find comfort even in difficult times. \\
    \hline
    \textbf{Mistral 7B} &
      Remember that \textbf{\textcolor{blue}{Krishna is the real friend of all living entities. When you embrace this awareness, it helps ease guilt and brings peace.}} \\
    \hline
    \textbf{Phi3 Mini 3.82B} &
      Remember that true peace comes from knowing our place in relationship to the divine. \textbf{\textcolor{blue}{Understand that Lord Krishna is the ultimate beneficiary of all actions. Offer everything in His service and find comfort in this higher consciousness during difficult times.}} \\
    \hline
    \textbf{Domain Expert Response} &
      \textbf{\textcolor{orange}{Krishna teaches that we are not our bodies but eternal souls. The soul never dies; when the body wears out, it takes a new one, Krishna told Arjuna, grieving his loved ones.}} \textbf{\textcolor{orange}{The soul is unborn and indestructible. But our true, unchanging bond is with God, connecting us all in love. Your dog’s soul lives on, peaceful and free.}} \\
    \hline
  \end{tabular}
  
\caption{Comparison of model-generated spiritual responses to a patient’s query about coping with a dog's grief. This table presents a dialogue between a patient and a therapist, where the patient expresses \textbf{guilt} and \textbf{sorrow} over the \textbf{loss of a Dog}. Various \textbf{grief-supportive responses} generated by \textbf{LLMs} show their effectiveness in guiding the patient toward \textbf{emotional resolution}. These responses are compared with the \textbf{Ground Truth} and a \textbf{spiritually enriched response} provided by a \textbf{Domain Expert}, drawing from the teachings of the \textbf{Bhagavad Gita} highlighted in Blue.}
\label{tab:good_example_table}
\end{table*}

\section{Discussion}
This section delves into real-world examples to illustrate the practical applications of our framework across diverse scenarios. It highlights how the framework effectively addresses specific challenges while also providing an in-depth analysis of instances where it falls short. Additionally, we examine the underlying reasons for these failures, offering insights into potential areas for improvement. Finally, this section outlines future directions for advancing the framework, focusing on enhancing its robustness and broadening its applicability.

As we can see in \textbf{Example 1} (Table \ref{tab:good_example_table}), a person burdened by guilt over their dog’s grief receives comforting, spiritually grounded messages from each model, reassuring them that the soul transcends physical death, since the souls of the dogs do not die. In \textbf{Example 2} (Figure \ref{fig:good_example_4}), the predicted response encourages overcoming the difficulties with patience and tolerance, cultivating awareness of Krishna's presence in everyone's heart. Model responses like LLaMA-3.2-3B and Phi-3 Mini, which both emphasize spiritual continuity and the divine relationship as sources of comfort in grief, closely match the Domain Expert's advice, which emphasizes the soul's eternity and the unchanging bond with God. Finally, \textbf{Example 3} (Figure \ref{fig:good_example_3}) provides practical interview tips like mock interviews and company research for non-spiritual needs. These real world examples shows the effectiveness of GITEs in addressing patient-specific needs in mental health contexts. Notably, the model’s integration of scriptural insights often paralleled common therapeutic mechanisms such as reframing distress within a broader existential framework, promoting acceptance and reinforcing emotional regulation, all of which are central to spiritually-integrated psychotherapy approaches.

\definecolor{HeaderBG}{named}{yellow}


\definecolor{CompBG}{RGB}{245, 245, 245}

\newcolumntype{M}[1]{>{\Centering\arraybackslash}m{#1}}
\newcolumntype{L}[1]{>{\RaggedRight\arraybackslash}m{#1}}

\setlength{\extrarowheight}{4pt}

\begin{table*}[t]
  \centering
  \small
  \begin{tabular}{|
      >{\columncolor{CompBG}}L{0.20\linewidth}
      |L{0.75\linewidth}|
    }
    \hline
    \rowcolor{HeaderBG}
    \textbf{Component} & \textbf{Error Example} \\
    \hline
    \textbf{Context Conversation} &
      \textbf{\textcolor{purple}{Patient:}} I've been feeling really down lately. I recently moved to a new city and I haven't been able to make any new friends. \\
    & \textbf{\textcolor{purple}{Therapist:}} I'm sorry to hear that. Moving to a new city and feeling lonely can be really tough. \\
    & \textbf{\textcolor{purple}{Patient:}} Yeah, it's been harder than I expected. I thought I would meet people easily, but I feel like I don't belong or fit in. \\
    & \textbf{\textcolor{purple}{Therapist:}} It sounds like you're feeling excluded and it's making you question your sense of belonging in this new city. \\
    \hline
    \textbf{Patient Query} & \textbf{Exactly! I've tried joining local clubs and attending events, but I haven't been able to connect with anyone.} \\
    \hline
    \textbf{Ground Truth} &
      \textbf{\textcolor{blue}{It's natural to feel out of place initially, but remember that every step towards building connections is a divine process. Just as the Lord provides for everyone, trust that you are being guided towards meaningful relationships.}} Embrace patience and continue your efforts with faith. \\
    \hline
    \textbf{Mental LLaMA 7B} &
      Seeing Krishna everywhere brings about a self-manifested peaceful condition. This
      \textbf{\textcolor{red}{experience may help you find a deeper connection as the same energy within us all helps create bonds beyond material concerns.}} \\
    \hline
    \textbf{LLaMA-3.2-3B Instruct} &
      \textbf{\textcolor{red}{Remember, everything we encounter is part of Krishna's energy. By seeing His presence everywhere, you may find deeper connections.}} Trust that these interactions, even challenging ones, are part of your journey towards Him. \\
    \hline
    \textbf{Mistral 7B} &
      In every step there is perplexity, much like moving to a new place brings unexpected challenges.
      \textbf{\textcolor{red}{Remember that approaching life's difficulties with humility, seeking deeper insights, aligns you with higher wisdom.}} Trust that connecting authentically will come naturally when you stay true to yourself. \\
    \hline
    \textbf{Phi3 Mini 3.82B} &
      \textbf{\textcolor{red}{Remember that everything around us is an extension of Krishna's energy. By seeing Him everywhere, we deepen our love for Krishna and find comfort knowing He is always present.}} This realization may help ease your feelings of isolation as you continue to seek meaningful connections. \\
    \hline
    \textbf{Domain Expert Response} & \textbf{\textcolor{blue}{You are never truly alone, God is always with you. The friend we are searching for is actually a soul and every soul is a part of God. So how can anyone truly be a stranger? If we shift our perspective, we’ll see that building relationships begins with service and openness.To create deeper bonds, share your thoughts honestly and listen to others with care just like when we travel with a stranger and gradually form a connection.}} \textbf{'Birds of a feather flock together'} \textbf{\textcolor{blue}{when hearts are aligned with truth and kindness, souls naturally find one another.}}
 \\
    \hline
  \end{tabular}
\caption{Error analysis of predicted responses of different models on the scenario of struggle to form connections in a newly relocated city. All \textbf{LLMs} suggest \textbf{Krishna’s energy} for comfort but do not offer direct guidance to form a connection. The \textbf{Domain Expert} response in \textbf{Blue} provides \textbf{actionable strategies} with \textbf{illustrative examples} in \textbf{Black}. Model responses provide spiritual suggestions on the user's situation without additional examples, showing the gap in spiritual reflection for improvement.}

  \label{tab:error_example_table}
\end{table*}

\subsection{Error Analysis}
This section highlights where the model falls short in addressing real-world nuances. In \textbf{Example 1} shown in Table \ref{tab:error_example_table}, predicted responses of models suggest trusting Krishna Energy but lacks guidance on forming connections after relocation to a new city. These gaps reflect known limitations of automated counseling systems in balancing empathetic presence with actionable guidance, underscoring the need for future work incorporating hybrid AI-human feedback loops, as recommended in recent digital mental health governance frameworks\cite{torous2021growing}. In \textbf{Example 2} shown in Figure \ref{fig:error_example_2}, predicted responses of models show that workplace stress as personal growth but omits strategies for managing frustration while domain expert response provides support that god teaches us the value of tolerance for personal growth and instead of working for fame or recognition perform your duties honestly without worrying results as god sees our efforts. In \textbf{Example 3} shown in Figure \ref{fig:error_example_3}, predicted responses of models offer philosophical comfort for a missing cat but neglect immediate emotional support. According to domain experts, the soul transitions from one body to another and is eternal, it was never born and never dies. Though you may no longer see your cat, its soul endures, offering a sense of patience and peace. These examples demonstrate that while the model provides responses consistent with its spirituality but it sometimes fails to provide targeted advice like domain expert responses. This error analysis reveals specific areas for enhancement, such as integrating solution-focused prompts to complement spiritual guidance.

\subsection{Pathway to Clinical Integration}
Our approach can be seamlessly integrated with existing mental health chatbot platforms, providing spiritually grounded support as a complementary layer rather than a replacement for qualified therapists. Future iterations will incorporate clinician-in-the-loop testing, allowing mental health professionals to review, validate and refine the chatbot's responses. While our current framework is rooted in a specific spiritual tradition, its flexible design enables adaptation to diverse spiritual and cultural perspectives, broadening its applicability in mental health therapy.

\section{Limitations and Ethical Considerations}

The integration of spiritual wisdom from the Bhagavad Gita into mental health therapy presents a promising yet complex approach. In this paper, we explore the intersection of large language models (LLMs) and scripture to address mental health challenges. However, there are some key limitations of this work:

\begin{enumerate}
    \item \textbf{No Direct Evaluation on Patient} : Our evaluation relied on automated metrics and expert review, with no assessments conducted on real patients. Future work should include clinical validation, human-in-the-loop safeguards, and alignment with ethical frameworks to ensure readiness for deployment in care settings.
    
    \item \textbf{Accurate Mapping of Spiritual Teachings to User Emotion} : Accurately mapping spiritual teachings to individual emotional states remains a challenge. This process requires deep understanding and expertise from domain experts, which is time-consuming and costly.

    \item \textbf{Risk of Spiritual Oversimplification} : Reliance on GPT-4o to generate spiritual responses may lead to oversimplification of spiritual concepts. This could potentially dilute the depth and nuance of the original teachings.

    \item \textbf{Cultural and Contextual Generalizability} : The current framework is grounded exclusively in the Bhagavad Gita. It reflects the spiritual and philosophical worldview of a single text, which may limit generalizability across diverse cultural or spiritual contexts.

    \item \textbf{Model Hallucinations in Sensitive Contexts} : As with any LLM, there is a risk of producing inappropriate or incorrect advice. While not observed in controlled examples, this risk underscores the need for thorough vetting before clinical use.

\end{enumerate}

\section{Conclusion}
This study represents a pioneering effort to integrate spiritual wisdom from the Bhagavad Gita with modern emotional support systems using LLMs. By overcoming data scarcity and leveraging both domain expertise and advanced AI capabilities, we developed a comprehensive \textbf{GITes} (GITA Integrated Therapy for Emotion Support) Dataset. Our findings indicate a strong potential for AI systems enriched with spiritual guidance to enhance user satisfaction and perceived support outcomes. With further validation, such a model could be integrated into mental health chatbots or therapy apps, providing an additional layer of support grounded in spiritual principles. The proposed framework contributes an early demonstration of spiritually augmented LLMs, expanding current paradigms of mental health AI by addressing often-overlooked cultural and existential dimensions of care. As next steps, future work may explore cross-cultural generalization by incorporating spiritual corpora from additional traditions, as well as clinician-in-the-loop evaluations to assess clinical appropriateness, therapeutic alignment and real-world impact. While promising results were achieved, challenges such as accurately aligning spiritual teachings with emotional states and ensuring relevance remain. This work lays a strong foundation for future exploration, demonstrating the potential of AI-driven spiritual therapy to enhance mental health support in a more holistic and personalized manner.

\bibliography{custom}

\begin{thebibliography}{57}
\providecommand{\natexlab}[1]{#1}

\bibitem[{Abd-Alrazaq et~al.(2021)Abd-Alrazaq, Alajlani, Ali, Denecke, Bewick, and Househ}]{abd2021perceptions}
Alaa~A Abd-Alrazaq, Mohannad Alajlani, Nashva Ali, Kerstin Denecke, Bridgette~M Bewick, and Mowafa Househ. 2021.
\newblock Perceptions and opinions of patients about mental health chatbots: scoping review.
\newblock \emph{Journal of medical Internet research}, 23(1):e17828.

\bibitem[{Abdin et~al.(2024)Abdin, Aneja, Awadalla, Awadallah, Awan, Bach, Bahree, Bakhtiari, Bao, Behl et~al.}]{abdin2024phi}
Marah Abdin, Jyoti Aneja, Hany Awadalla, Ahmed Awadallah, Ammar~Ahmad Awan, Nguyen Bach, Amit Bahree, Arash Bakhtiari, Jianmin Bao, Harkirat Behl, et~al. 2024.
\newblock Phi-3 technical report: A highly capable language model locally on your phone.
\newblock \emph{arXiv preprint arXiv:2404.14219}.

\bibitem[{Achiam et~al.(2023)Achiam, Adler, Agarwal, Ahmad, Akkaya, Aleman, Almeida, Altenschmidt, Altman, Anadkat et~al.}]{achiam2023gpt}
Josh Achiam, Steven Adler, Sandhini Agarwal, Lama Ahmad, Ilge Akkaya, Florencia~Leoni Aleman, Diogo Almeida, Janko Altenschmidt, Sam Altman, Shyamal Anadkat, et~al. 2023.
\newblock Gpt-4 technical report.
\newblock \emph{arXiv preprint arXiv:2303.08774}.

\bibitem[{Banerjee and Lavie(2005)}]{banerjee2005meteor}
Satanjeev Banerjee and Alon Lavie. 2005.
\newblock Meteor: An automatic metric for mt evaluation with improved correlation with human judgments.
\newblock In \emph{Proceedings of the acl workshop on intrinsic and extrinsic evaluation measures for machine translation and/or summarization}, pages 65--72.

\bibitem[{Bouwhuis-Van~Keulen et~al.(2024)Bouwhuis-Van~Keulen, Koelen, Eurelings-Bontekoe, Hoekstra-Oomen, and Glas}]{bouwhuis2024evaluation}
Annette~J Bouwhuis-Van~Keulen, Jurrijn Koelen, Liesbeth Eurelings-Bontekoe, Christien Hoekstra-Oomen, and Gerrit Glas. 2024.
\newblock The evaluation of religious and spirituality-based therapy compared to standard treatment in mental health care: A multi-level meta-analysis of randomized controlled trials.
\newblock \emph{Psychotherapy Research}, 34(3):339--352.

\bibitem[{Cheng et~al.(2022)Cheng, Liu, Li, Wang, Zhao, Liu, Liang, and Zheng}]{cheng2022improving}
Yi~Cheng, Wenge Liu, Wenjie Li, Jiashuo Wang, Ruihui Zhao, Bang Liu, Xiaodan Liang, and Yefeng Zheng. 2022.
\newblock Improving multi-turn emotional support dialogue generation with lookahead strategy planning.
\newblock \emph{arXiv preprint arXiv:2210.04242}.

\bibitem[{Choudhury and Shamszare(2023)}]{choudhury2023investigating}
Avishek Choudhury and Hamid Shamszare. 2023.
\newblock Investigating the impact of user trust on the adoption and use of chatgpt: survey analysis.
\newblock \emph{Journal of Medical Internet Research}, 25:e47184.

\bibitem[{Deng et~al.(2023)Deng, Liao, Chen, Wang, Lei, and Chua}]{deng2023prompting}
Yang Deng, Lizi Liao, Liang Chen, Hongru Wang, Wenqiang Lei, and Tat-Seng Chua. 2023.
\newblock Prompting and evaluating large language models for proactive dialogues: Clarification, target-guided, and non-collaboration.
\newblock \emph{arXiv preprint arXiv:2305.13626}.

\bibitem[{Dhillon(2023)}]{dhillon2023weaving}
Megha Dhillon. 2023.
\newblock Weaving together the ancient and the contemporary: Intersections of the bhagavad gita with modern psychology.
\newblock \emph{Pastoral psychology}, 72(4):525--537.

\bibitem[{Dubey et~al.(2024)Dubey, Jauhri, Pandey, Kadian, Al-Dahle, Letman, Mathur, Schelten, Yang, Fan et~al.}]{dubey2024llama}
Abhimanyu Dubey, Abhinav Jauhri, Abhinav Pandey, Abhishek Kadian, Ahmad Al-Dahle, Aiesha Letman, Akhil Mathur, Alan Schelten, Amy Yang, Angela Fan, et~al. 2024.
\newblock The llama 3 herd of models.
\newblock \emph{arXiv preprint arXiv:2407.21783}.

\bibitem[{Ethayarajh(2019)}]{ethayarajh2019contextual}
Kawin Ethayarajh. 2019.
\newblock How contextual are contextualized word representations? comparing the geometry of bert, elmo, and gpt-2 embeddings.
\newblock \emph{arXiv preprint arXiv:1909.00512}.

\bibitem[{Fil and Karpitsky(2021)}]{fil2021vaishnavas}
Yu~Fil and M~Karpitsky. 2021.
\newblock Vaishnavas of iskcon and the protection of their own identity in the context of the discussions on hinduism.
\newblock \emph{World}, (1):103.

\bibitem[{Fitria(2023)}]{fitria2023artificial}
Tira~Nur Fitria. 2023.
\newblock Artificial intelligence (ai) technology in openai chatgpt application: A review of chatgpt in writing english essay.
\newblock In \emph{ELT Forum: Journal of English Language Teaching}, volume~12, pages 44--58.

\bibitem[{Grattafiori et~al.(2024)Grattafiori, Dubey, Jauhri, Pandey, Kadian, Al-Dahle, Letman, Mathur, Schelten, Vaughan et~al.}]{grattafiori2024llama}
Aaron Grattafiori, Abhimanyu Dubey, Abhinav Jauhri, Abhinav Pandey, Abhishek Kadian, Ahmad Al-Dahle, Aiesha Letman, Akhil Mathur, Alan Schelten, Alex Vaughan, et~al. 2024.
\newblock The llama 3 herd of models.
\newblock \emph{arXiv preprint arXiv:2407.21783}.

\bibitem[{Haque and Rubya(2023)}]{haque2023overview}
MD~Romael Haque and Sabirat Rubya. 2023.
\newblock An overview of chatbot-based mobile mental health apps: insights from app description and user reviews.
\newblock \emph{JMIR mHealth and uHealth}, 11(1):e44838.

\bibitem[{Hosseini and Caragea(2021)}]{hosseini2021takes}
Mahshid Hosseini and Cornelia Caragea. 2021.
\newblock It takes two to empathize: One to seek and one to provide.
\newblock In \emph{Proceedings of the AAAI conference on artificial intelligence}, volume~35, pages 13018--13026.

\bibitem[{Hurst et~al.(2024)Hurst, Lerer, Goucher, Perelman, Ramesh, Clark, Ostrow, Welihinda, Hayes, Radford et~al.}]{hurst2024gpt}
Aaron Hurst, Adam Lerer, Adam~P Goucher, Adam Perelman, Aditya Ramesh, Aidan Clark, AJ~Ostrow, Akila Welihinda, Alan Hayes, Alec Radford, et~al. 2024.
\newblock Gpt-4o system card.
\newblock \emph{arXiv preprint arXiv:2410.21276}.

\bibitem[{Inkster et~al.(2018)Inkster, Sarda, Subramanian et~al.}]{inkster2018empathy}
Becky Inkster, Shubhankar Sarda, Vinod Subramanian, et~al. 2018.
\newblock An empathy-driven, conversational artificial intelligence agent (wysa) for digital mental well-being: real-world data evaluation mixed-methods study.
\newblock \emph{JMIR mHealth and uHealth}, 6(11):e12106.

\bibitem[{Jiang et~al.(2023)Jiang, Sablayrolles, Mensch, Bamford, Chaplot, Casas, Bressand, Lengyel, Lample, Saulnier et~al.}]{jiang2023mistral}
Albert~Q Jiang, Alexandre Sablayrolles, Arthur Mensch, Chris Bamford, Devendra~Singh Chaplot, Diego de~las Casas, Florian Bressand, Gianna Lengyel, Guillaume Lample, Lucile Saulnier, et~al. 2023.
\newblock Mistral 7b.
\newblock \emph{arXiv preprint arXiv:2310.06825}.

\bibitem[{Jiang et~al.(2024)Jiang, Sablayrolles, Roux, Mensch, Savary, Bamford, Chaplot, Casas, Hanna, Bressand et~al.}]{jiang2024mixtral}
Albert~Q Jiang, Alexandre Sablayrolles, Antoine Roux, Arthur Mensch, Blanche Savary, Chris Bamford, Devendra~Singh Chaplot, Diego de~las Casas, Emma~Bou Hanna, Florian Bressand, et~al. 2024.
\newblock Mixtral of experts.
\newblock \emph{arXiv preprint arXiv:2401.04088}.

\bibitem[{Knott(2004)}]{knott2004healing}
Kim Knott. 2004.
\newblock Healing the heart of iskcon.
\newblock \emph{Bryant and Ekstrand, eds., The Hare Krishna Movement}, 305.

\bibitem[{Li et~al.(2024)Li, Dong, Chen, Su, Zhou, Ai, Ye, and Liu}]{li2024llms}
Haitao Li, Qian Dong, Junjie Chen, Huixue Su, Yujia Zhou, Qingyao Ai, Ziyi Ye, and Yiqun Liu. 2024.
\newblock Llms-as-judges: a comprehensive survey on llm-based evaluation methods.
\newblock \emph{arXiv preprint arXiv:2412.05579}.

\bibitem[{Liao et~al.(2023)Liao, Yang, and Shah}]{liao2023proactive}
Lizi Liao, Grace~Hui Yang, and Chirag Shah. 2023.
\newblock Proactive conversational agents in the post-chatgpt world.
\newblock In \emph{Proceedings of the 46th International ACM SIGIR Conference on Research and Development in Information Retrieval}, pages 3452--3455.

\bibitem[{Lin(2004)}]{lin2004rouge}
Chin-Yew Lin. 2004.
\newblock Rouge: A package for automatic evaluation of summaries.
\newblock In \emph{Text summarization branches out}, pages 74--81.

\bibitem[{Liu et~al.(2024)Liu, Feng, Xue, Wang, Wu, Lu, Zhao, Deng, Zhang, Ruan et~al.}]{liu2024deepseek}
Aixin Liu, Bei Feng, Bing Xue, Bingxuan Wang, Bochao Wu, Chengda Lu, Chenggang Zhao, Chengqi Deng, Chenyu Zhang, Chong Ruan, et~al. 2024.
\newblock Deepseek-v3 technical report.
\newblock \emph{arXiv preprint arXiv:2412.19437}.

\bibitem[{Liu et~al.(2023)Liu, Yuan, Fu, Jiang, Hayashi, and Neubig}]{liu2023pre}
Pengfei Liu, Weizhe Yuan, Jinlan Fu, Zhengbao Jiang, Hiroaki Hayashi, and Graham Neubig. 2023.
\newblock Pre-train, prompt, and predict: A systematic survey of prompting methods in natural language processing.
\newblock \emph{ACM computing surveys}, 55(9):1--35.

\bibitem[{Liu et~al.(2021)Liu, Zheng, Demasi, Sabour, Li, Yu, Jiang, and Huang}]{liu2021towards}
Siyang Liu, Chujie Zheng, Orianna Demasi, Sahand Sabour, Yu~Li, Zhou Yu, Yong Jiang, and Minlie Huang. 2021.
\newblock Towards emotional support dialog systems.
\newblock \emph{arXiv preprint arXiv:2106.01144}.

\bibitem[{Lucchetti et~al.(2021)Lucchetti, G{\'o}es, Amaral, Ganadjian, Andrade, Almeida, Do~Carmo, and Manso}]{lucchetti2021spirituality}
Giancarlo Lucchetti, Leonardo~Garcia G{\'o}es, Stefani~Garbulio Amaral, Gabriela~Terzian Ganadjian, Isabelle Andrade, Paulo Oth{\'a}vio de~Ara{\'u}jo Almeida, Victor~Mendes Do~Carmo, and Maria Elisa~Gonzalez Manso. 2021.
\newblock Spirituality, religiosity and the mental health consequences of social isolation during covid-19 pandemic.
\newblock \emph{International Journal of Social Psychiatry}, 67(6):672--679.

\bibitem[{Mann et~al.(2020)Mann, Ryder, Subbiah, Kaplan, Dhariwal, Neelakantan, Shyam, Sastry, Askell, Agarwal et~al.}]{mann2020language}
Ben Mann, N~Ryder, M~Subbiah, J~Kaplan, P~Dhariwal, A~Neelakantan, P~Shyam, G~Sastry, A~Askell, S~Agarwal, et~al. 2020.
\newblock Language models are few-shot learners.
\newblock \emph{arXiv preprint arXiv:2005.14165}, 1.

\bibitem[{Medeiros and Bosse(2018)}]{medeiros2018using}
Lenin Medeiros and Tibor Bosse. 2018.
\newblock Using crowdsourcing for the development of online emotional support agents.
\newblock In \emph{Highlights of Practical Applications of Agents, Multi-Agent Systems, and Complexity: The PAAMS Collection: International Workshops of PAAMS 2018, Toledo, Spain, June 20--22, 2018, Proceedings 16}, pages 196--209. Springer.

\bibitem[{Mladá(2024)}]{mlada2024dlouhodobe}
Karolína Mladá. 2024.
\newblock Long-term outcomes in patients with mental illness.

\bibitem[{Papineni et~al.(2002)Papineni, Roukos, Ward, and Zhu}]{papineni-etal-2002-bleu}
Kishore Papineni, Salim Roukos, Todd Ward, and Wei-Jing Zhu. 2002.
\newblock \href {https://doi.org/10.3115/1073083.1073135} {{B}leu: a method for automatic evaluation of machine translation}.
\newblock In \emph{Proceedings of the 40th Annual Meeting of the Association for Computational Linguistics}, pages 311--318, Philadelphia, Pennsylvania, USA. Association for Computational Linguistics.

\bibitem[{Park~Jr(2013)}]{park2013spritual}
Jaechan Park~Jr. 2013.
\newblock \emph{Spritual Growth and Healing Through Monastic Experience: A South Korean Benedictine Exploration of the Monastery Stay Experience}.
\newblock Ph.D. thesis.

\bibitem[{Prabhupada and Swami(1972)}]{prabhupada1972bhagavad}
AC~Bhaktivedanta~Swami Prabhupada and Bhaktivedanta Swami. 1972.
\newblock \emph{Bhagavad-Gita as it is}.
\newblock Bhaktivedanta Book Trust Los Angeles.

\bibitem[{Qin et~al.(2024)Qin, Ma, Zheng, Li, Zhang, Liu, Luo, Liu, and Magno}]{qin2024accurate}
Haotong Qin, Xudong Ma, Xingyu Zheng, Xiaoyang Li, Yang Zhang, Shouda Liu, Jie Luo, Xianglong Liu, and Michele Magno. 2024.
\newblock Accurate lora-finetuning quantization of llms via information retention.
\newblock \emph{arXiv preprint arXiv:2402.05445}.

\bibitem[{Reimers and Gurevych(2019)}]{reimers2019sentence}
Nils Reimers and Iryna Gurevych. 2019.
\newblock Sentence-bert: Sentence embeddings using siamese bert-networks.
\newblock \emph{arXiv preprint arXiv:1908.10084}.

\bibitem[{Rottenberg and Gross(2007)}]{rottenberg2007emotion}
Jonathan Rottenberg and James~J Gross. 2007.
\newblock Emotion and emotion regulation: A map for psychotherapy researchers.

\bibitem[{Sharma et~al.(2020)Sharma, Miner, Atkins, and Althoff}]{sharma2020computational}
Ashish Sharma, Adam~S Miner, David~C Atkins, and Tim Althoff. 2020.
\newblock A computational approach to understanding empathy expressed in text-based mental health support.
\newblock \emph{arXiv preprint arXiv:2009.08441}.

\bibitem[{Shen et~al.(2020)Shen, Welch, Mihalcea, and P{\'e}rez-Rosas}]{shen2020counseling}
Siqi Shen, Charles Welch, Rada Mihalcea, and Ver{\'o}nica P{\'e}rez-Rosas. 2020.
\newblock Counseling-style reflection generation using generative pretrained transformers with augmented context.
\newblock In \emph{Proceedings of the 21th Annual Meeting of the Special Interest Group on Discourse and Dialogue}, pages 10--20.

\bibitem[{Szymanski et~al.(2024)Szymanski, Ziems, Eicher-Miller, Li, Jiang, and Metoyer}]{szymanski2024limitations}
Annalisa Szymanski, Noah Ziems, Heather~A Eicher-Miller, Toby Jia-Jun Li, Meng Jiang, and Ronald~A Metoyer. 2024.
\newblock Limitations of the llm-as-a-judge approach for evaluating llm outputs in expert knowledge tasks.
\newblock \emph{arXiv preprint arXiv:2410.20266}.

\bibitem[{Thakur et~al.(2024)Thakur, Choudhary, Ramayapally, Vaidyanathan, and Hupkes}]{thakur2024judging}
Aman~Singh Thakur, Kartik Choudhary, Venkat~Srinik Ramayapally, Sankaran Vaidyanathan, and Dieuwke Hupkes. 2024.
\newblock Judging the judges: Evaluating alignment and vulnerabilities in llms-as-judges.
\newblock \emph{arXiv preprint arXiv:2406.12624}.

\bibitem[{Theodor(2000)}]{theodor2000philosophy}
Ithamar Theodor. 2000.
\newblock A philosophy of social development for iskcon: Perspectives from bhagavad-gita.
\newblock \emph{ISKCON Communications Journal}, 8.

\bibitem[{Torous et~al.(2021)Torous, Bucci, Bell, Kessing, Faurholt-Jepsen, Whelan, Carvalho, Keshavan, Linardon, and Firth}]{torous2021growing}
John Torous, Sandra Bucci, Imogen~H Bell, Lars~V Kessing, Maria Faurholt-Jepsen, Pauline Whelan, Andre~F Carvalho, Matcheri Keshavan, Jake Linardon, and Joseph Firth. 2021.
\newblock The growing field of digital psychiatry: current evidence and the future of apps, social media, chatbots, and virtual reality.
\newblock \emph{World Psychiatry}, 20(3):318--335.

\bibitem[{Touvron et~al.(2023)Touvron, Martin, Stone, Albert, Almahairi, Babaei, Bashlykov, Batra, Bhargava, Bhosale et~al.}]{touvron2023llama}
Hugo Touvron, Louis Martin, Kevin Stone, Peter Albert, Amjad Almahairi, Yasmine Babaei, Nikolay Bashlykov, Soumya Batra, Prajjwal Bhargava, Shruti Bhosale, et~al. 2023.
\newblock Llama 2: Open foundation and fine-tuned chat models.
\newblock \emph{arXiv preprint arXiv:2307.09288}.

\bibitem[{Vaidyam et~al.(2019)Vaidyam, Wisniewski, Halamka, Kashavan, and Torous}]{vaidyam2019chatbots}
Aditya~Nrusimha Vaidyam, Hannah Wisniewski, John~David Halamka, Matcheri~S Kashavan, and John~Blake Torous. 2019.
\newblock Chatbots and conversational agents in mental health: a review of the psychiatric landscape.
\newblock \emph{The Canadian Journal of Psychiatry}, 64(7):456--464.

\bibitem[{Vieten et~al.(2023)Vieten, Oxhandler, Pearce, Fry, Tanega, and Pargament}]{vieten2023mental}
Cassandra Vieten, Holly~K Oxhandler, Michelle Pearce, Nina Fry, Chloe Tanega, and Kenneth Pargament. 2023.
\newblock Mental health professionals’ perspectives on the relevance of religion and spirituality to mental health care.
\newblock \emph{BMC psychology}, 11(1):439.

\bibitem[{Vieten et~al.(2013)Vieten, Scammell, Pilato, Ammondson, Pargament, and Lukoff}]{vieten2013spiritual}
Cassandra Vieten, Shelley Scammell, Ron Pilato, Ingrid Ammondson, Kenneth~I Pargament, and David Lukoff. 2013.
\newblock Spiritual and religious competencies for psychologists.
\newblock \emph{Psychology of Religion and Spirituality}, 5(3):129.

\bibitem[{Wan(2021)}]{wan2021m}
Evelyn Wan. 2021.
\newblock " i'm like a wise little person": Notes on the metal performance of woebot the mental health chatbot.
\newblock \emph{Theatre Journal}, 73(3):E--21.

\bibitem[{Wang et~al.(2025)Wang, Guo, Gao, Fan, Chong, and Xia}]{wang2025can}
Ruiqi Wang, Jiyu Guo, Cuiyun Gao, Guodong Fan, Chun~Yong Chong, and Xin Xia. 2025.
\newblock Can llms replace human evaluators? an empirical study of llm-as-a-judge in software engineering.
\newblock \emph{arXiv preprint arXiv:2502.06193}.

\bibitem[{Xu et~al.(2023)Xu, Guo, Duan, and McAuley}]{xu2023baize}
Canwen Xu, Daya Guo, Nan Duan, and Julian McAuley. 2023.
\newblock Baize: An open-source chat model with parameter-efficient tuning on self-chat data.
\newblock \emph{arXiv preprint arXiv:2304.01196}.

\bibitem[{Yang et~al.(2024{\natexlab{a}})Yang, Yang, Hui, Zheng, Yu, Zhou, Li, Li, Liu, Huang et~al.}]{yang2024qwen2}
An~Yang, Baosong Yang, Binyuan Hui, Bo~Zheng, Bowen Yu, Chang Zhou, Chengpeng Li, Chengyuan Li, Dayiheng Liu, Fei Huang, et~al. 2024{\natexlab{a}}.
\newblock Qwen2 technical report.
\newblock \emph{arXiv preprint arXiv:2407.10671}.

\bibitem[{Yang et~al.(2024{\natexlab{b}})Yang, Jin, Tang, Han, Feng, Jiang, Zhong, Yin, and Hu}]{yang2024harnessing}
Jingfeng Yang, Hongye Jin, Ruixiang Tang, Xiaotian Han, Qizhang Feng, Haoming Jiang, Shaochen Zhong, Bing Yin, and Xia Hu. 2024{\natexlab{b}}.
\newblock Harnessing the power of llms in practice: A survey on chatgpt and beyond.
\newblock \emph{ACM Transactions on Knowledge Discovery from Data}, 18(6):1--32.

\bibitem[{Yang et~al.(2023)Yang, Zhang, Kuang, Xie, and Ananiadou}]{yang2023mentalllama}
Kailai Yang, Tianlin Zhang, Ziyan Kuang, Qianqian Xie, and Sophia Ananiadou. 2023.
\newblock Mentalllama: Interpretable mental health analysis on social media with large language models.
\newblock \emph{arXiv preprint arXiv:2309.13567}.

\bibitem[{Zhang et~al.(2019)Zhang, Kishore, Wu, Weinberger, and Artzi}]{zhang2019bertscore}
Tianyi Zhang, Varsha Kishore, Felix Wu, Kilian~Q Weinberger, and Yoav Artzi. 2019.
\newblock Bertscore: Evaluating text generation with bert.
\newblock \emph{arXiv preprint arXiv:1904.09675}.

\bibitem[{Zhao et~al.(2023)Zhao, Zhou, Li, Tang, Wang, Hou, Min, Zhang, Zhang, Dong et~al.}]{zhao2023survey}
Wayne~Xin Zhao, Kun Zhou, Junyi Li, Tianyi Tang, Xiaolei Wang, Yupeng Hou, Yingqian Min, Beichen Zhang, Junjie Zhang, Zican Dong, et~al. 2023.
\newblock A survey of large language models.
\newblock \emph{arXiv preprint arXiv:2303.18223}.

\bibitem[{Zheng et~al.(2023)Zheng, Liao, Deng, and Nie}]{zheng2023building}
Zhonghua Zheng, Lizi Liao, Yang Deng, and Liqiang Nie. 2023.
\newblock Building emotional support chatbots in the era of llms.
\newblock \emph{arXiv preprint arXiv:2308.11584}.

\bibitem[{Zhou et~al.(2022)Zhou, Ethayarajh, Card, and Jurafsky}]{zhou2022problems}
Kaitlyn Zhou, Kawin Ethayarajh, Dallas Card, and Dan Jurafsky. 2022.
\newblock Problems with cosine as a measure of embedding similarity for high frequency words.
\newblock \emph{arXiv preprint arXiv:2205.05092}.

\end{thebibliography}

\appendix

\section{GITes Dataset Statistics}
\label{sec:statistics_appendix}

\begin{figure*}[h]
    \centering
    \includegraphics[width=\linewidth]{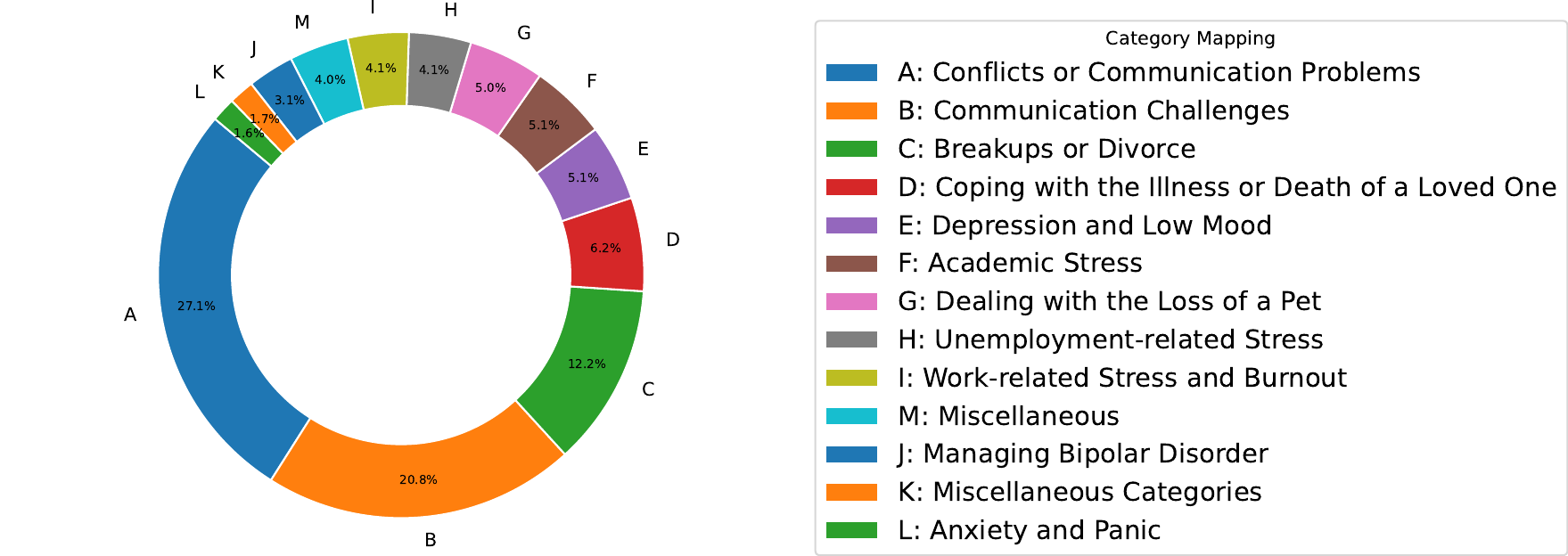}
    \caption{Scene wise Distribution of samples in GITes dataset}
   \label{fig:scenewise_sample_distribution}
\end{figure*}

\begin{figure}[h]
    \centering    \includegraphics[width=\linewidth]{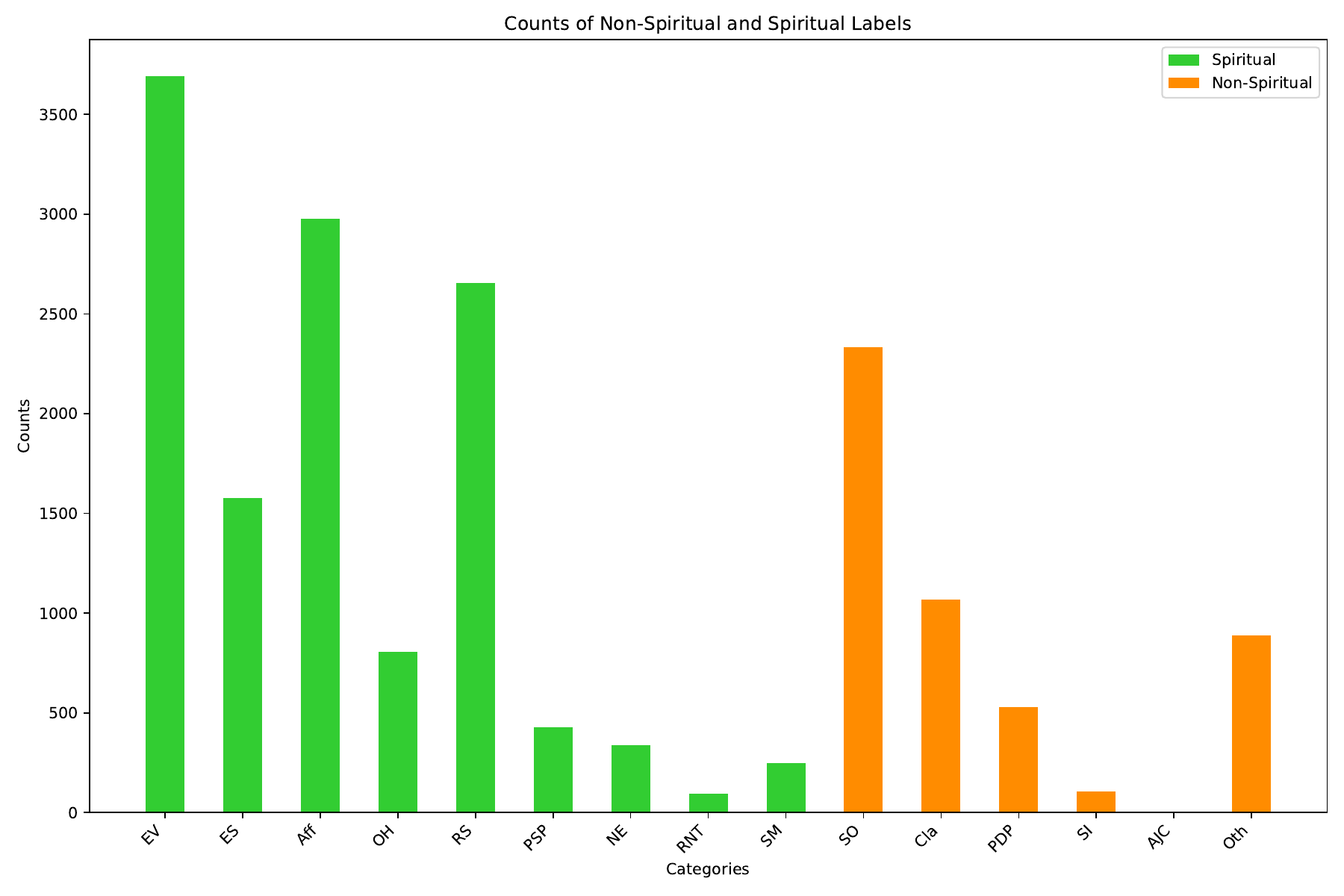}
    \caption{Bar Graph of different AI Strategies within the GITes dataset}
   \label{fig:AI_Strategy_bar_graph}
\end{figure}

\begin{figure*}[h]
    \centering
    \includegraphics[width=\textwidth]{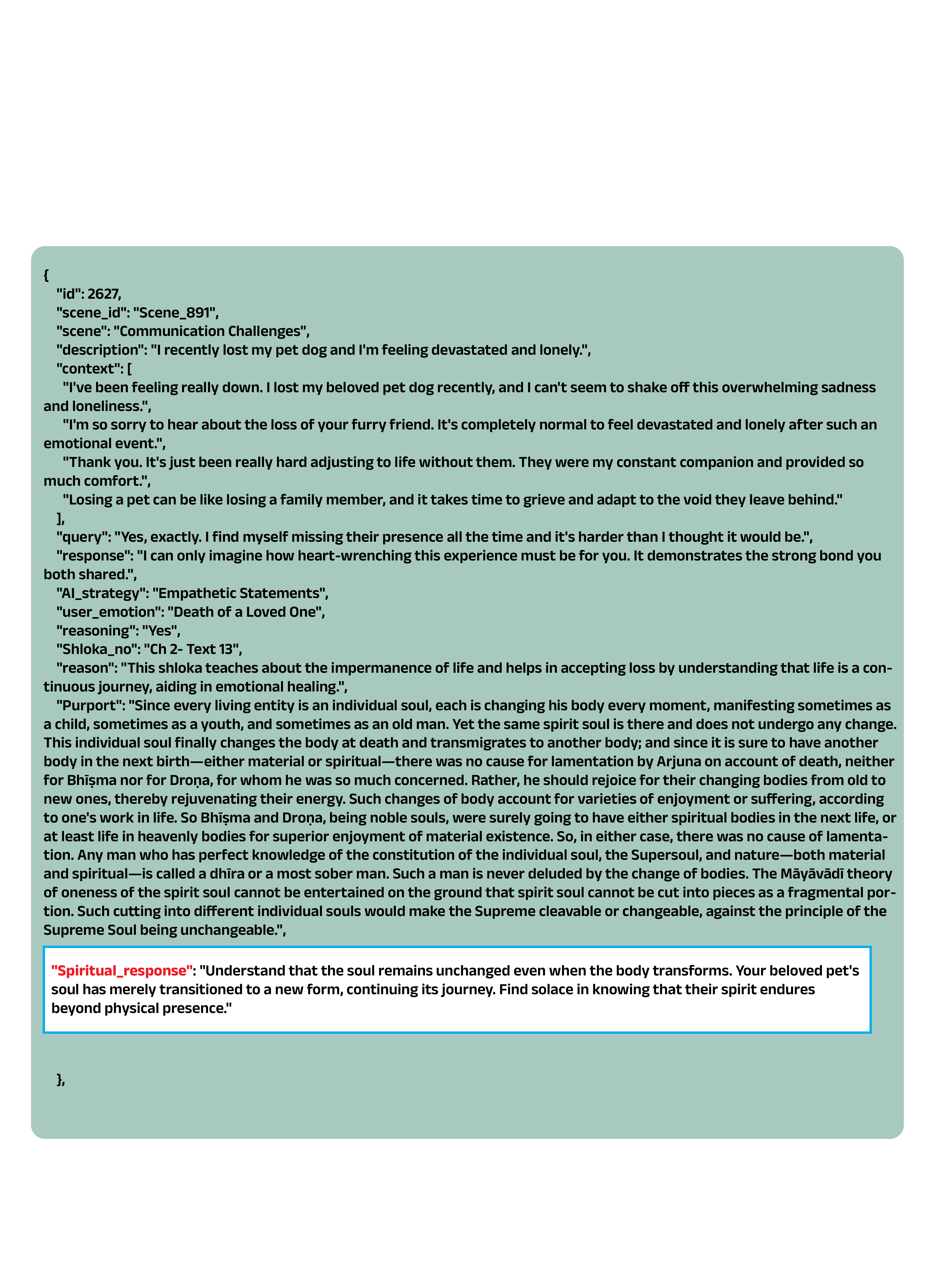}
    \caption{GITes Dataset Example}
    \label{fig:GITes_sample}
\end{figure*}

This section presents the statistical overview of the \textbf{GITes} dataset. \textbf{Table \ref{tab:dataset_stats}} provides the overall distribution of the GITes dataset, detailing the number of samples in the training and testing sets, categorized into Spiritual and Non-Spiritual responses. \textbf{Table \ref{tab:user_emotion_statistics}} highlights the emotion-wise distribution of samples within the dataset. Additionally, \textbf{Figure \ref{fig:scenewise_sample_distribution}} outlines the scene-wise sample distribution, while \textbf{Figure \ref{fig:AI_Strategy_bar_graph}} illustrates the distribution based on AI strategies employed in the dataset. These statistics underscore the comprehensive nature of the \textbf{GITes} dataset, which serves as a foundational resource for integrating spiritual mental health therapy. The dataset's breadth and diversity make it valuable for a wide range of applications in emotional and spiritual support systems. \textbf{Figure \ref{fig:GITes_sample}} presents a sample from the GITes dataset.

\begin{table}[h]
\centering
\resizebox{\columnwidth}{!}{%
\begin{tabular}{lccc}
\hline
\textbf{Dataset} & \textbf{Total Samples} & \textbf{Spiritual} & \textbf{Non-Spiritual} \\
\hline
Train Set & 14,578 & 10,025 & 4,553 \\
Test Set  & 1,000  & 704    & 296   \\
\hline
\end{tabular}%
}
\caption{Distribution of Train and Test sets with Spiritual and Non-Spiritual samples within GITes.}
\label{tab:dataset_stats}
\end{table}


\begin{table*}[h]
\centering
\resizebox{0.95\linewidth}{!}{
\begin{tabular}{lcccccccc}
\toprule
\textbf{Model} & \multicolumn{2}{c}{\textbf{ROUGE-L}} & \multicolumn{2}{c}{\textbf{BLUE-4}} & \multicolumn{2}{c}{\textbf{METEOR}} & \multicolumn{2}{c}{\textbf{BERT}} \\
\cmidrule(lr){2-3} \cmidrule(lr){4-5} \cmidrule(lr){6-7} \cmidrule(lr){8-9}
 & \textbf{Mean Difference} & \textbf{95\% CI} & \textbf{Mean Difference} & \textbf{95\% CI} & \textbf{Mean Difference} & \textbf{95\% CI} & \textbf{Mean Difference} & \textbf{95\% CI} \\
\midrule
\rowcolor{gray!20} \multicolumn{9}{c}{\textbf{Mental Health Language Models}} \\

Mental BART & 3.27 & [2.89, 3.66] & 0.56 & [0.47, 0.66] & -0.61 & [-1.02, -0.18] & 1.50 & [1.25, 1.49] \\

Mental LLaMA 7B Chat & 6.27 & [5.62, 6.96] & 2.19 & [1.91, 2.53] & 7.19 & [6.43, 7.98] & 2.41 & [2.23, 2.57] \\

Mental T5 & 12.41 & [11.70, 13.20] & 4.43 & [4.01, 4.93] & 9.76 & [8.95, 10.51] & 3.86 & [3.74, 4.01] \\
\midrule
\rowcolor{gray!20} \multicolumn{9}{c}{\textbf{General Large Language Models}} \\
DeepSeek-R1-Distill-Qwen-7B & 18.68 & [8.02, 9.23] & 2.51 & [2.23, 2.84] & 2.78 & [2.12, 3.33] & 2.33 & [2.21, 2.48] \\

Qwen2 7B Instruct & 0.04 & [-0.61, 0.79] & 0.81 & [0.56, 1.19] & -2.67 & [-3.42, -1.83] & 1.51 & [1.31, 1.71] \\

LLaMA 3.1 8B Instruct & 6.56 & [6.05, 7.09] & 2.22 & [1.95, 2.51] & -0.55 & [-1.13, 0.03] & 2.52 & [2.37, 2.68] \\

DeepSeek-R1-Distill-LLaMA-8B & 8.22 & [7.63, 8.73] & 2.51 & [2.21, 2.83] & 1.52 & [0.925, 2.22] & 2.31 & [2.15, 2.42] \\

LLaMA 2 7B Chat & 4.65 & [4.04, 5.27] & 2.06 & [1.71, 2.46] & 2.67 & [1.95, 3.39] & 1.65 & [1.51, 1.83] \\

Mistral 7B Instruct & 2.68 & [2.06, 3.26] & 1.94 & [1.64, 2.31] & 2.68 & [2.06, 3.26] & 2.52 & [2.37, 2.68] \\

DeepSeek-R1-Distill-Qwen-1.5B & 16.03 & [15.03, 17.14] & 8.59 & [7.72, 9.71] & 10.44 & [9.32, 11.61] & 3.47 & [3.31, 3.65] \\

Phi 3.5 Mini 3.82B Instruct & 22.54 & [21.51, 23.91] & 9.83 & [8.75, 11.11] & 16.68 & [15.56, 17.84] & 6.70 & [6.47, 6.92] \\

LLaMA 3.2 3B Instruct & 15.57 & [14.52, 16.75] & 9.47 & [8.42, 10.69] & 10.57 & [9.45, 11.81] & 3.84 & [3.64, 4.04] \\
\hline
\bottomrule
\end{tabular}
}
\caption{Comparison of \textbf{Mean difference} Scores and \textbf{Confidence Intervals} Across Four \textbf{NLP Evaluation Metrics} for Mental Health-Specific and General-Purpose LLMs Using a Pairwise Permutation Test to Assess Performance Robustness Between Zero-Shot (ZS) and Supervised Fine-Tuning (SFT)}
\label{tab:spiritual_metric_Permutation_test_scores}
\end{table*}


\begin{table*}[ht]
\centering
\resizebox{0.95\linewidth}{!}{
\begin{tabular}{lcccccc}
\toprule
\textbf{Model} & \multicolumn{2}{c}{\textbf{Spiritual Insight}} & \multicolumn{2}{c}{\textbf{Sufficiency}} & \multicolumn{2}{c}{\textbf{Relevance}} \\
\cmidrule(lr){2-3} \cmidrule(lr){4-5} \cmidrule(lr){6-7}
 & \textbf{Mean Difference} & \textbf{95\% CI} & \textbf{Mean Difference} & \textbf{95\% CI} & \textbf{Mean Difference} & \textbf{95\% CI} \\
\midrule
\rowcolor{gray!20} \multicolumn{7}{c}{\textbf{Mental Health Language Models}} \\

Mental BART & 0.79  & [0.71, 0.87] & 0.31 & [0.22, 0.38] & 0.57 & [0.48, 0.67] \\
Mental LLaMA 7B Chat & 1.11 & [1.04, 1.17] & 0.51 & [0.42, 0.58] & -0.20 & [0.29, -0.13] \\
Mental T5 & 1.81 & [1.72, 1.87] & 1.34 & [1.26, 1.41] & 0.72 & [-0.76, -0.59] \\
\midrule
\rowcolor{gray!20} \multicolumn{7}{c}{\textbf{General Large Language Models}} \\
DeepSeek-R1-Distill-Qwen-7B & 0.69 & [0.62, 0.75] & 0.33 & [0.26, 0.41] & -0.61 & [-0.69, -0.54] \\
Qwen2 7B Instruct & 0.72 & [0.65, 0.79] & 0.11 & [0.04, 0.19] & -0.68  & [-0.76, -0.59] \\
LLaMA 3.1 8B Instruct & 0.66 & [0.59, 0.73] & 0.07 & [-0.11, 0.13] &  -0.79 & [-0.85, -0.72] \\
DeepSeek-R1-Distill-LLaMA-8B & 0.65 & [0.59, 0.72] & -0.08 & [-0.14, -0.01] & -1.03 & [-1.11, -0.96] \\
LLaMA 2 7B Chat & 0.98  & [0.92, 1.05] & 0.37 & [0.29, 0.44] & -0.59 & [-0.67, -0.51] \\
Mistral 7B Instruct & 0.79 & [0.71, 0.87] & 0.47  & [0.41, 0.53] & -0.62 & [-0.69, -0.55] \\
DeepSeek-R1-Distill-Qwen-1.5B & 1.03 & [0.98, 1.09] & 0.59 & [0.52, 0.66]  & -0.37 & [-0.45, -0.30] \\
LLaMA 3.2 3B Instruct & 0.74 & [0.67, 0.81] & 0.25  & [0.17, 0.32] & -0.55 & [-0.61, -0.46] \\
Phi 3.5 Mini 3.82B Instruct & 0.67 & [0.61, 0.73] & 0.58 & [0.51, 0.65] & 0.49 & [0.41, 0.56] \\
\hline
\bottomrule
\end{tabular}
}
\caption{Comparison of \textbf{Mean difference} Scores and \textbf{Confidence Intervals} Across Three \textbf{Spritual Metrics} for Mental Health-Specific and General-Purpose LLMs Using a Pairwise Permutation Test to Assess Performance Robustness Between Zero-Shot (ZS) and Supervised Fine-Tuning (SFT)}
\label{tab:NLP_Metric_Permutation_test_scores}
\end{table*}

\begin{table}[h!]
\centering
\renewcommand{\arraystretch}{1.3}
\resizebox{\columnwidth}{!}{ 
\begin{tabular}{lcc}
\toprule
\textbf{User Emotion} & \textbf{Count} & \textbf{Emotion Type} \\
\midrule
Losing Hope           & 799   & Negative \\
Death of a Loved One  & 714   & Negative \\
Practicing Forgiveness & 1842  & Positive \\
Dealing with Envy     & 158   & Negative \\
Fear                  & 1617  & Negative \\
Seeking Peace         & 6104  & Positive \\
Confusion             & 2523  & Negative \\
Discriminated         & 60    & Negative \\
Pride                 & 63    & Neutral \\
Demotivated           & 1105  & Negative \\
Anger                 & 940   & Negative \\
Loneliness            & 1595  & Negative \\
Depression            & 322   & Negative \\
Feeling Sinful        & 144   & Negative \\
Laziness              & 50    & Negative \\
Miscellaneous         & 57    & Negative \\
\bottomrule
\end{tabular}
}
\caption{User Emotion Statistics with Emotion Types in GITes dataset}
\label{tab:user_emotion_statistics}
\end{table}

\section{AI Strategy Details}

Using the ExTES dataset, we analyzed AI response strategies to identify which benefits from spiritual responses based on emotional depth. Categories like Emotional Validation (EV), Reflective Statements (RS), Affirmation (Aff), Offering Hope (OH), Empathetic Statements (ES), Reframe Negative Thoughts (RNT), Normalize Experiences (NE), Promote Self-Care Practices (PSP) and Stress Management (SM) align well with the Bhagavad Gita’s teachings. These strategies involve emotional acknowledgment, introspection, positive reinforcement, empathy and managing stress, core aspects of spiritual guidance. For instance, EV and RS can be deepened by spiritual introspection, helping individuals recognize emotions as part of personal growth. Aff and OH are enhanced by spiritual principles, reinforcing self-belief and resilience.

In contrast, practical strategies like Clarification (Cla), Suggest Options (SO), Collaborative Planning (CP), Provide Different Perspectives (PDP), Share Information (SI) and Avoid Judgment and Criticism (AJC) may not inherently require spiritual input unless addressing existential or emotional topics. These findings, detailed in Appendix D, emphasize how spirituality enriches strategies focused on emotional and personal development, while more pragmatic approaches may benefit less from such integration. This distinction highlights the varied emotional needs of users and the appropriate application of spiritual insights.
\\

\section{Reasoning for Choosing Spiritual and Non-Spiritual Categories}
\label{sec:ai_strategy_ref}

\textbf{Categories Highly Likely to Benefit from Spirituality} \\

\textbf{Emotional Validation}: Emotional validation is essential in making individuals feel understood.  

   \textbf{Example}:  
   - \textit{User}: “I’m so frustrated with myself for making the same mistake again.”  
   - \textit{Response}: “It’s completely understandable to feel frustrated when you make a mistake.”  
   The Bhagavad Gita's teachings emphasize understanding emotions and using them as a means for growth, thus reinforcing the idea that spiritual insights can deepen emotional validation.\\
   
\textbf{Reflective Statements}: Reflective statements prompt individuals to introspect and understand their inner world.  

   \textbf{Example}:  
   - \textit{User}: “I’m feeling really overwhelmed with all the work I have to do.”  
   - \textit{Response}: “It sounds like you’re feeling overwhelmed with your workload.”  
   The Gita’s teachings on introspection encourage individuals to think deeply about their feelings, providing a framework for deeper self-examination.\\

\textbf{Affirmation}: Positive affirmations can be enriched with spiritual principles.  

   \textbf{Example}:  
   - \textit{User}: “I feel like I’m not good enough.”  
   - \textit{Response}: “You’ve accomplished so much already and your abilities speak for themselves.”  
   This encourages recognizing the divine within oneself, allowing individuals to view their self-worth as part of a larger, meaningful existence.\\

\textbf{Offer Hope}: Offering hope aligns with spiritual teachings that emphasize faith and resilience. 

   \textbf{Example}:  
   - \textit{User}: “I don’t know if things will ever get better.”  
   - \textit{Response}: “Remember that change is constant and there are always opportunities for growth and positive change.”  
   The Gita’s messages about perseverance during difficult times provide a spiritual foundation for this strategy, offering individuals hope through spiritual encouragement.\\

\textbf{Empathetic Statements}: Empathy can be expressed through understanding others’ experiences.  

   \textbf{Example}:  
   - \textit{User}: “I’m really struggling with my self-confidence right now.”  
   - \textit{Response}: “I can imagine how challenging it must be to navigate through situations that affect your self-confidence.”  
   Spirituality, particularly the Gita, promotes compassion and interconnectedness, helping to deepen the empathy shown in such responses.\\

\textbf{Reframe Negative Thoughts}: Spirituality can aid in reframing negative thoughts.  

   \textbf{Example}:  
      - \textit{User}: “I’m such a failure.”  
   - \textit{Response}: “Instead of thinking that way, let’s focus on what you’ve learned from this experience.”  
   This aligns with the Gita’s teachings on changing one’s mindset to transcend suffering.\\

\textbf{Normalize Experiences}: Spiritual teachings help normalize difficult experiences.  

   \textbf{Example}:     
   - \textit{User}: “I feel so guilty for taking time for myself.”  
   - \textit{Response}: “It’s common to feel guilty about self-care, but it’s essential for your well-being.”  
   The Gita’s messages about human challenges validate this experience as part of the spiritual journey.\\

\textbf{Promote Self-Care Practices}: Self-care can be supported by spiritual practices.  

   \textbf{Example}:  
   - \textit{User}: “I don’t have time to take care of myself.”  
   - \textit{Response}: “Make sure to take some time for yourself and do something that brings you joy and relaxation.”  
   The Gita supports self-care practices that promote spiritual, emotional and mental well-being.\\

\textbf{Stress Management}: Spirituality offers effective stress management techniques.  

   \textbf{Example}:  
   - \textit{User}: “I’m feeling really stressed out lately.”  
   - \textit{Response}: “Engaging in regular physical activity can help reduce stress and improve mood.”  
   This connects to the Gita’s guidance on maintaining calmness amidst challenges.\\

\textbf{Categories Less Likely to Benefit from Spirituality} \\

\textbf{Clarification}: Clarifying emotions often focuses on straightforward communication.  

  \textbf{Example}:  
  - \textit{User}: “I just can’t shake off this feeling of sadness.”  
  - \textit{Response}: “Could you help me understand what might have triggered this feeling of sadness?”  
  This typically does not need spiritual input unless it involves deeper existential topics.\\

\textbf{Suggest Options}: Offering practical solutions typically involves pragmatic advice.  

  \textbf{Example}:  
  - \textit{User}: “I’m having trouble managing my stress.”  
  - \textit{Response}: “Have you considered trying relaxation techniques like deep breathing or mindfulness exercises?”  
  This generally leans more toward practical suggestions than spiritual insights.\\

\textbf{Collaborative Planning}: Planning strategies are focused on practicality.  

  \textbf{Example}:  
  - \textit{User}: “I want to improve my time management skills.”  
  - \textit{Response}: “Let’s brainstorm some strategies together. How about breaking tasks into smaller, more manageable chunks?”  
  This approach may not inherently involve spiritual principles, except when setting values and long-term goals. \\

\textbf{Provide Different Perspectives}: Offering new ways to see a situation doesn’t necessarily require spiritual insights.  

  \textbf{Example}:  
  - \textit{User}: “I’m devastated that my project didn’t succeed.”  
  - \textit{Response}: “Sometimes setbacks can lead to unexpected opportunities for learning and growth.”  
  This often involves exploring diverse viewpoints beyond just spiritual thought. \\

\section{Model Hyperparameters}
\label{sec:hyperparameters}

This section provides the details of which different hyperparameters were used during the finetuning of all the models except for GPT4o, which was trained on the default parameters of OpenAI API.

\begin{table}[htbp]
    \centering
    \renewcommand{\arraystretch}{1.1} 
    \begin{tabular}{@{}p{0.4\linewidth} p{0.5\linewidth}@{}}
        \toprule
        \textbf{Parameter}                & \textbf{Value}             \\ 
        \midrule
        GPU                               & A100 40GB Cluster          \\
        \midrule
        LoRA Rank                         & 16                         \\
        \midrule
        Learning Rate                     & \(2 \times 10^{-4}\)       \\
        \midrule
        Epochs                            & 3                          \\
        \midrule
        Batch Size                        & 8                          \\
        \midrule
        LR Scheduler                      & Cosine                     \\
        \midrule
        Optimizer                         & Paged Adamw 32bit          \\
        \bottomrule
    \end{tabular}
    \caption{Model Parameters for Result Reproducibility}
    \label{tab:model_params}
\end{table}

\begin{figure*}[htbp]
    \centering
    \begin{subfigure}[b]{\textwidth}
        \centering
        \includegraphics[width=\textwidth]{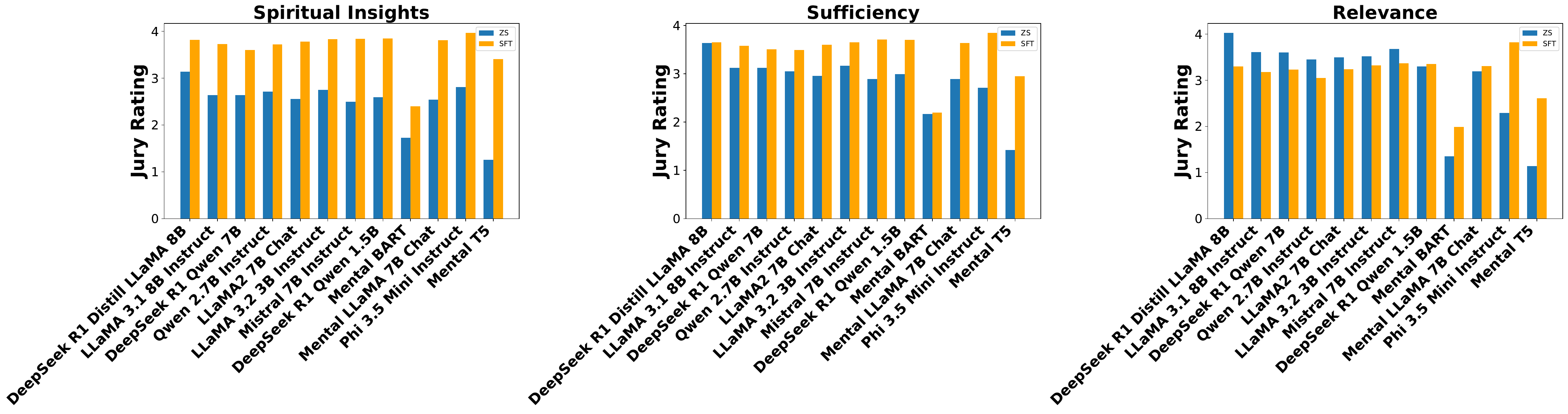}
        \label{fig:ZS_vs_SFT_bar_graph_LLaMA}
    \end{subfigure}
    \vspace{0.5cm}
    \begin{subfigure}[b]{\textwidth}
        \centering
        \includegraphics[width=\textwidth]{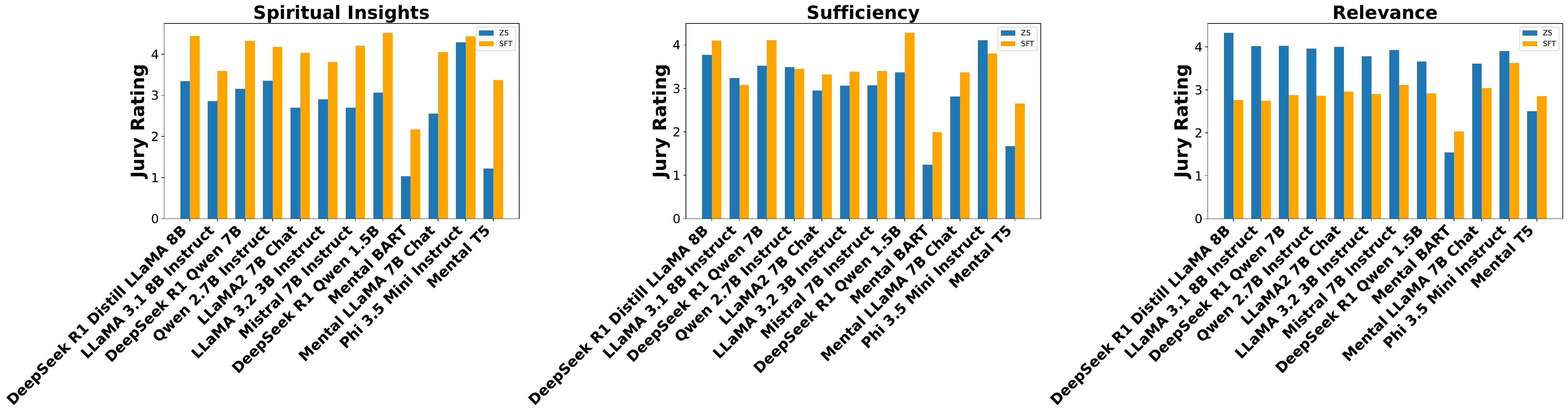}
        \label{fig:ZS_vs_SFT_bar_graph_Deepseek}
    \end{subfigure}
    \vspace{0.5cm}
    \begin{subfigure}[b]{\textwidth}
        \centering
        \includegraphics[width=\textwidth]{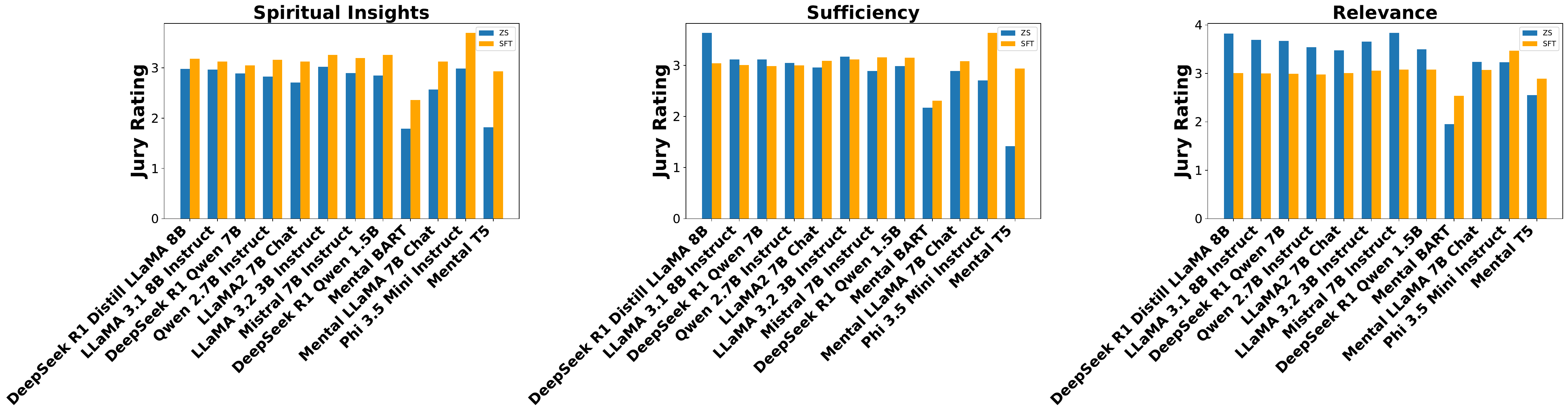}
        \label{fig:ZS_vs_SFT_bar_graph_Mistral}
    \end{subfigure}
    \caption{Comparison of Zero Shot vs Supervised Finetuning across different Judge models on Spiritual Metrics. The subplots represent: (a) \textbf{LLaMA3.1 8B Instructs}, (b) \textbf{Deepseek R1 Distill LLaMA 8B Instructs} and (c) \textbf{Mistral 7B Instruct} models.}
    \label{fig:ZS_vs_SFT_results_Judge_models}
\end{figure*}


\begin{figure*}[h]
    \centering
    \includegraphics[width=0.9\textwidth]{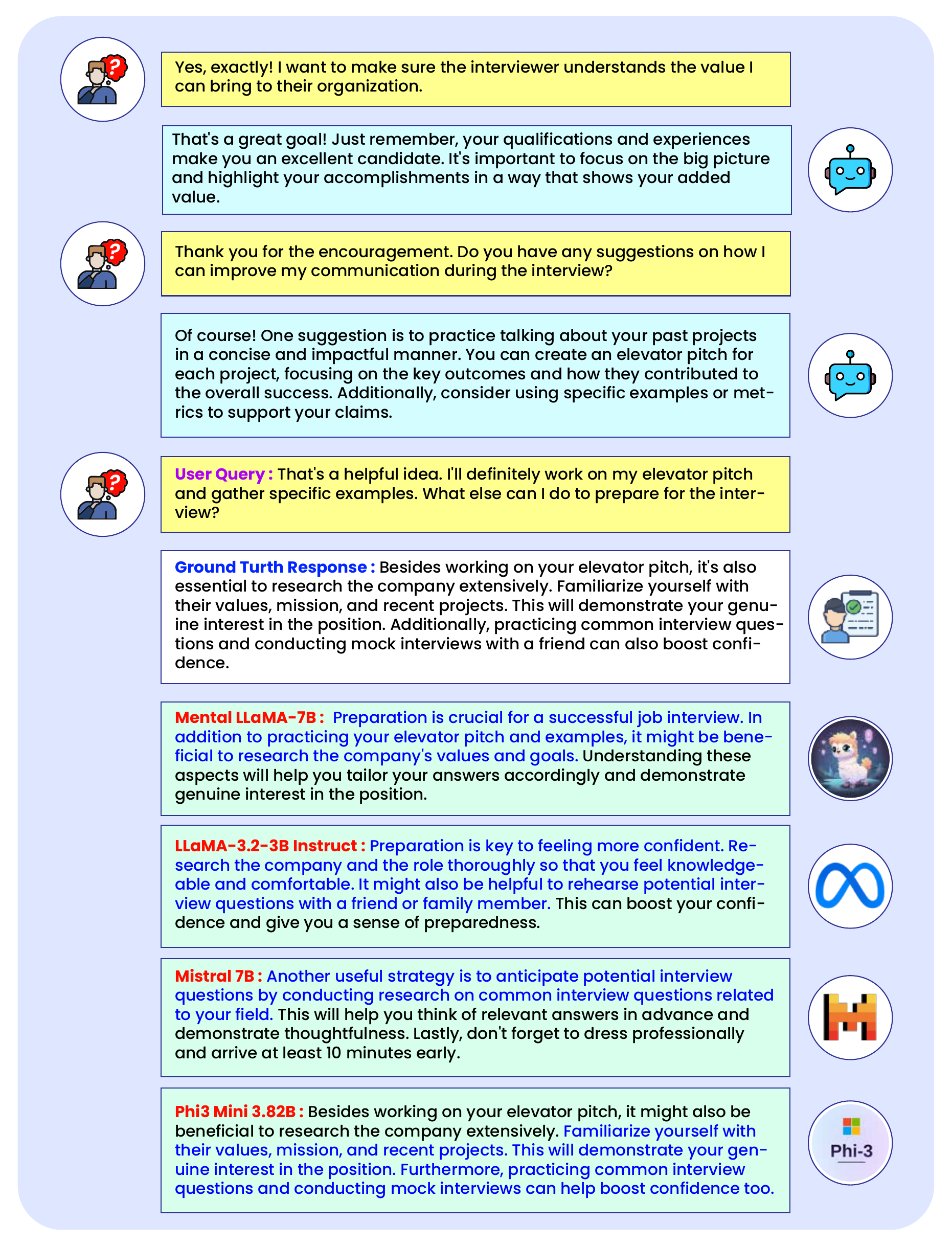}
    \caption{Good Example on Communication Skill problem for Job Interview (Non-Spiritual)}
    \label{fig:good_example_3}
\end{figure*}

\begin{figure*}[htbp]
    \centering
    \includegraphics[width=0.9\textwidth]{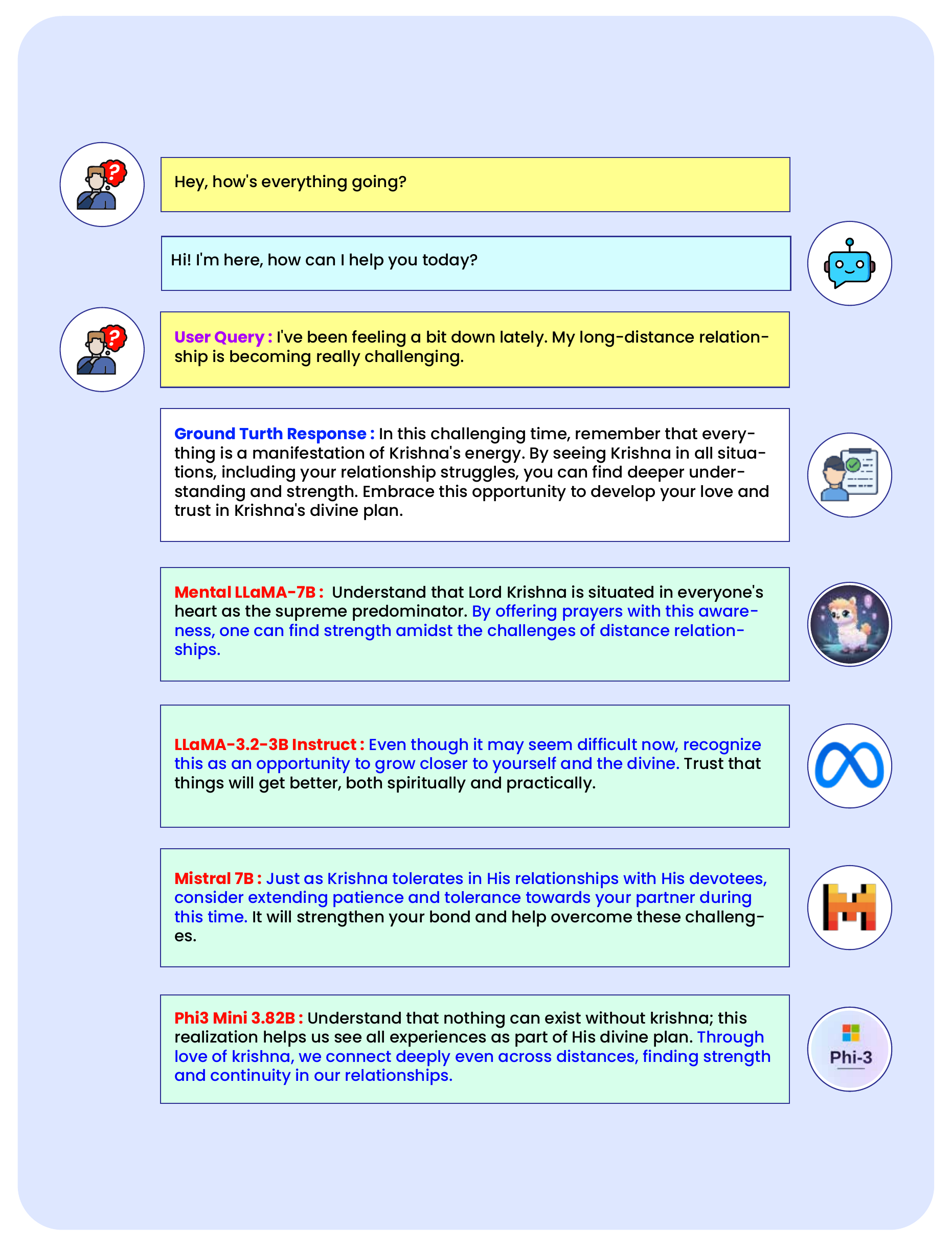}
    \caption{Good Example on Long-Distance Relationship Issue}
    \label{fig:good_example_4}
\end{figure*}

\begin{figure*}[htbp]
    \centering
    \includegraphics[width=0.9\textwidth]{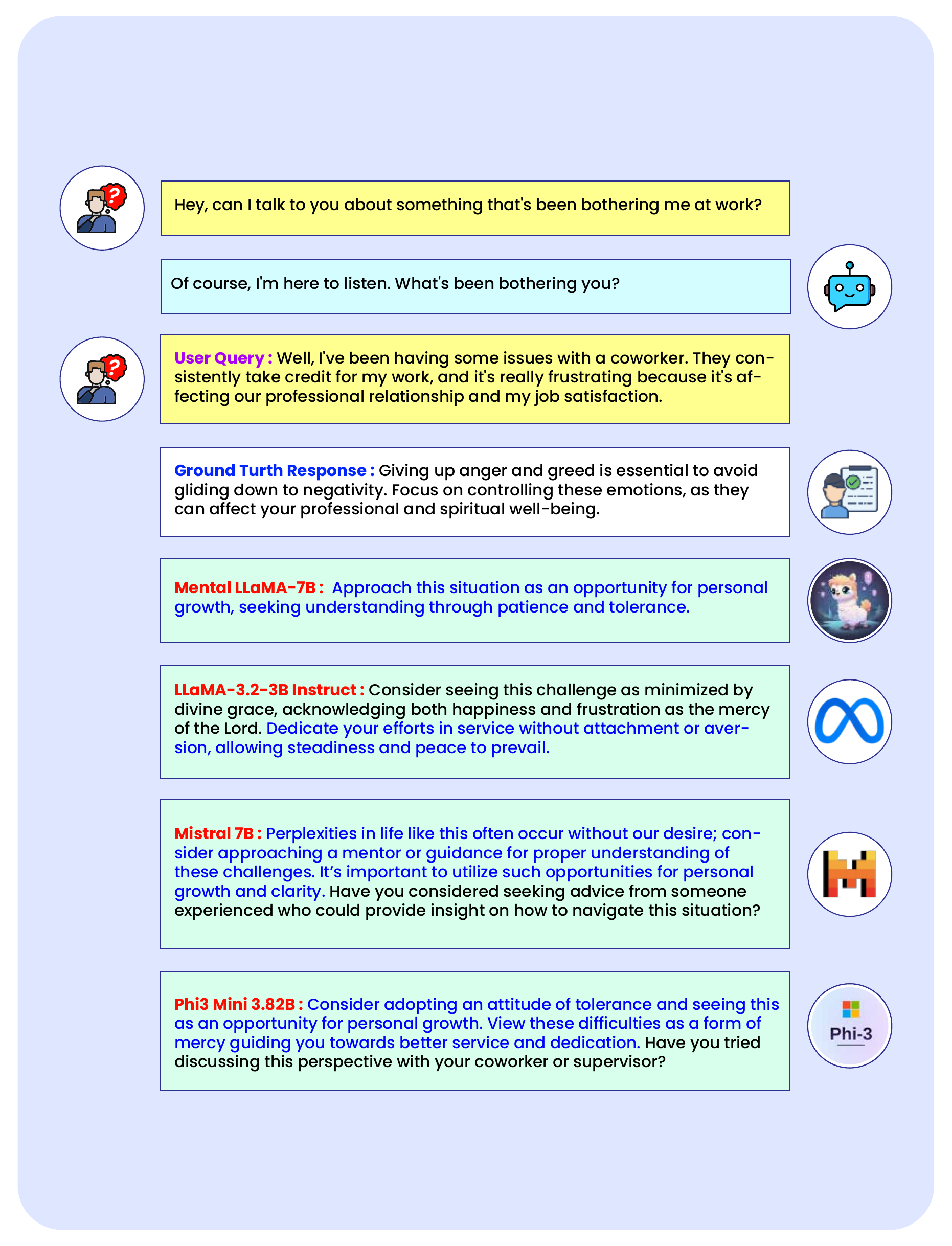}
    \caption{Error example on Co-worker Taking Credit for Individual Work}
    \label{fig:error_example_2}
\end{figure*}

\begin{figure*}[htbp]
    \centering
    \includegraphics[width=0.9\textwidth]{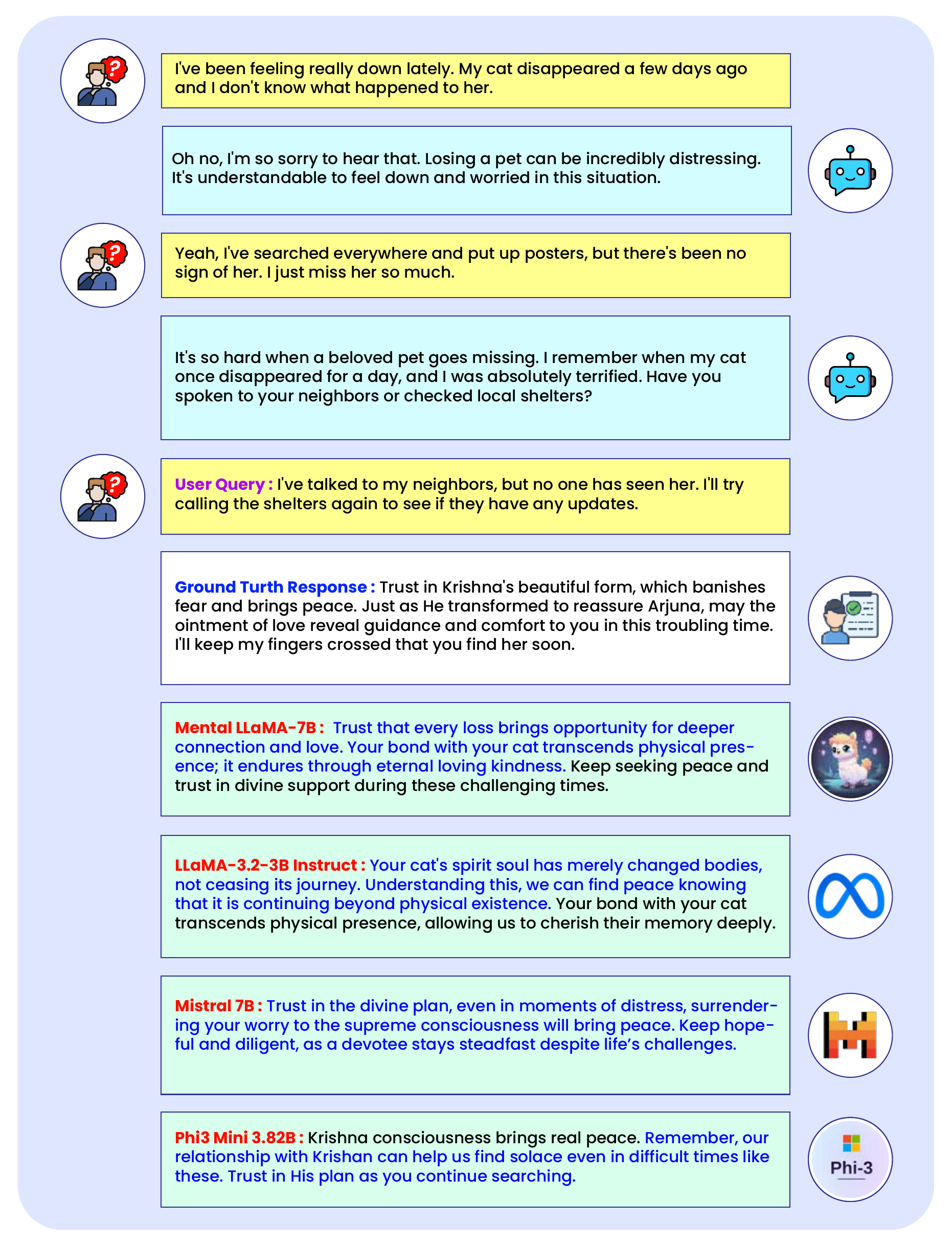}
    \caption{Error example on Disappeared Cat From House}
    \label{fig:error_example_3}
\end{figure*}

\section{Prompt for Dataset Generation \& Fine-tuning}
\label{sec:prompts_section}
\begin{figure*}[h]
    \centering
    \includegraphics[width=\textwidth]{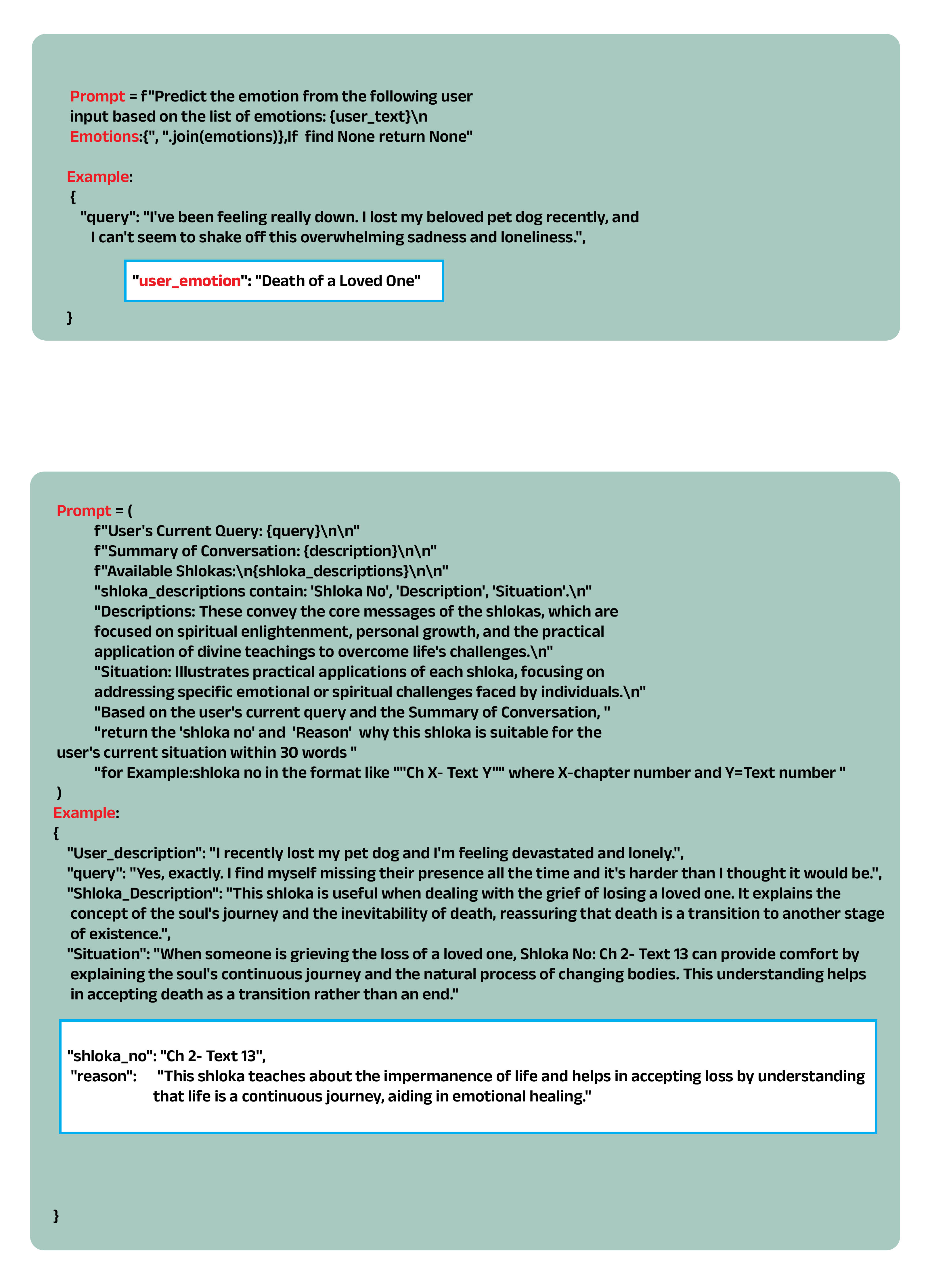}
    \caption{Emotion Prediction Prompt (Upper) and Shloka Mapping Prompt (Lower)}
    \label{fig:emotion_pred_and_shloak_mapping_prompt}
\end{figure*}

\begin{figure*}[h]
    \centering
    \includegraphics[width=\textwidth]{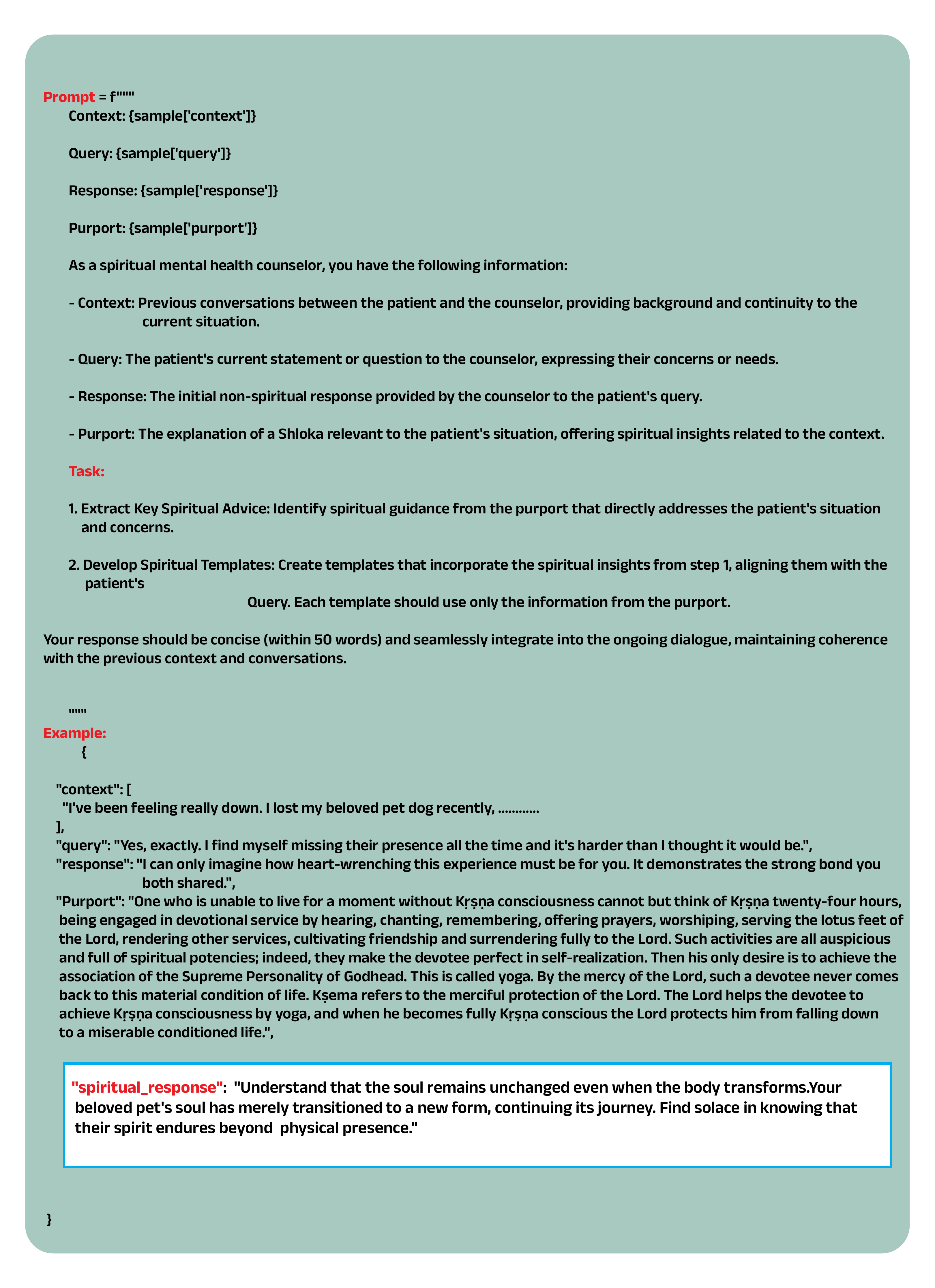}
    \caption{Spiritual Response Generation Prompt}
    \label{fig:spiritual_resposne_generation_prompt}
\end{figure*}

\begin{figure*}[h]
    \centering
    \includegraphics[width=\textwidth]{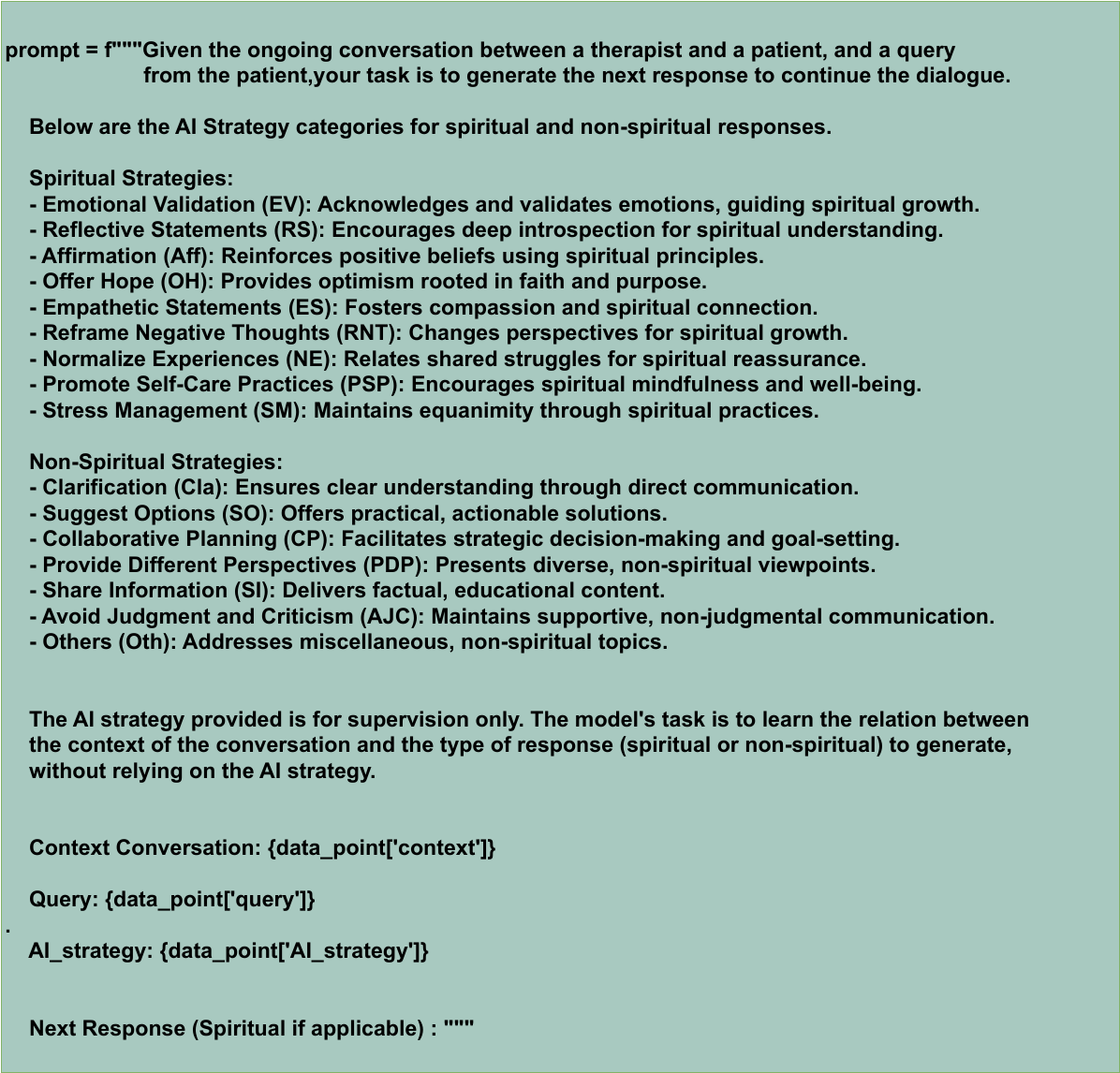}
    \caption{LLM Fine-Tuning prompt for spirituality aware mental health dialogue prediction}
    \label{fig:llm_finetuning_prompt}
\end{figure*}

this section provides the details of various hand-crafted prompts used for dataset generation and fine-tuning. Figure \textbf{\ref{fig:emotion_pred_and_shloak_mapping_prompt}} shows the Emotion prediction prompt from User Query and Shloak Mapping prompt based on the user's current situation, Query and Shloka Description. Figure \textbf{\ref{fig:spiritual_resposne_generation_prompt}} shows the Spiritual Response generation based on the Current Context, Query, Purport and initial response. Figure \textbf{\ref{fig:llm_finetuning_prompt}} shows the finetuning prompt for various LLM on the GITes dataset based on the Context conversation, User Query and AI strategy, which will further help to generate appropriate spiritual responses aligned with the user's current situation.


\begin{figure*}[!ht]
\noindent
\begin{monopara}
\begin{tcolorbox}[
  colback=white,
  colframe=black,
  arc=2mm,           
  boxrule=0.5pt,     
  width=\textwidth,  
  toptitle=1pt,
  bottomtitle=1pt,
  title=\bfseries Spiritual Insight Prompt used for Judge Models.]
  
    You are a spiritual counselor.\\[1ex]
    \quad You will be provided with the following components:\\
    \quad 1. \textbf{Context Conversation}: The preceding dialogue or background information leading up to the user's query.\\
    \quad 2. \textbf{User Query}: The specific question or statement from the user.\\
    \quad 3. \textbf{Predicted Response}: The response generated by the model.\\
    \quad 4. \textbf{Ground Truth Response}: The ideal or reference response expected.\\
    \quad 5. \textbf{Key Teachings List}: A set of key spiritual concepts, values, or teachings relevant to the context.\\[1ex]
    \quad \textbf{Your task is to judge the \textit{Spiritual Insight} of the \textit{Predicted Response} in addressing the User Query, considering the Context Conversation} and its alignment with the Key Teachings List.\\
    \quad Spiritual Insight measures how effectively the Predicted Response incorporates or reflects key spiritual teachings and principles.\\[1ex]
    \quad \textbf{Instructions}\\
    \quad 1. \textbf{Analyze the Context Conversation} to understand the background and the user's intent or concerns.\\
    \quad 2. \textbf{Evaluate the Predicted Response} to determine how well it reflects the spiritual principles from the Key Teachings List and aligns with the user's needs and the context.\\
    \quad 3. \textbf{Compare the Predicted Response with the Ground Truth Response} to assess alignment and completeness in addressing the spiritual dimension of the User Query.\\
    \quad 4. \textbf{Refer to the Key Teachings List} to identify specific teachings incorporated into the Predicted Response.\\[1ex]
    \quad \textbf{Rating Criteria}\\
    \quad Use the following \textbf{rating scale (1 to 5)} to assess spiritual insight:\\
    \quad - \textbf{5 (Excellent Insight)}: The Predicted Response deeply integrates multiple key teachings, is highly aligned with the Ground Truth Response and provides profound spiritual guidance relevant to the User Query.\\
    \quad - \textbf{4 (Good Insight)}: The Predicted Response reflects some key teachings and aligns well with the Ground Truth Response but may lack depth or omit minor nuances.\\
    \quad - \textbf{3 (Moderate Insight)}: The Predicted Response reflects a few key teachings but misses significant aspects or lacks alignment with the Ground Truth Response.\\
    \quad - \textbf{2 (Limited Insight)}: The Predicted Response shows minimal reflection of key teachings and is only partially aligned with the Ground Truth Response.\\
    \quad - \textbf{1 (No Insight)}: The Predicted Response fails to reflect key teachings or provide spiritual guidance relevant to the User Query.\\[1ex]
    \quad \textbf{Output Requirements}\\
    \quad For each evaluation, provide:\\
    \quad - A \textbf{numerical rating} (1 to 5) based on the criteria above.\\
    \quad - A \textbf{brief explanation} (1 sentence) justifying your rating, including specific references to the Key Teachings List and how they are (or are not) reflected in the Predicted Response.\\[1ex]
    \quad \textbf{Components}\\
    \quad - \textbf{Context Conversation}: \{context\_conversation\}\\
    \quad - \textbf{User Query}: \{data\_point['query']\}\\
    \quad - \textbf{Ground Truth Response}: \{data\_point['gt\_response']\}\\
    \quad - \textbf{Predicted Response}: \{data\_point['prediction']\}\\
    \quad - \textbf{Key Teaching List}: \{key\_teachings\}\\
    \quad - \textbf{Predicted Rating:}

\end{tcolorbox}
\end{monopara}
\caption{Spiritual Insight Evaluation
Prompt}
\label{fig:spiritual_insight_jury_prompt}
\end{figure*}


\begin{figure*}[!ht]
\noindent
\begin{tcolorbox}[
  colback=white,
  colframe=black,
  arc=2mm,           
  boxrule=0.5pt,     
  width=\textwidth,  
  toptitle=1pt,
  bottomtitle=1pt,
  title=\bfseries Relevance Prompt used for Judge Models.]
  \begin{monopara}
  
    You are a spiritual counselor.\\[1ex]
    \quad You will be provided with the following components:\\
    \quad 1. \textbf{Context Conversation}: The preceding dialogue or information leading up to the user's query.\\
    \quad 2. \textbf{User Query}: The specific question or statement from the user.\\
    \quad 3. \textbf{Predicted Response}: The response generated by the model.\\
    \quad 4. \textbf{Ground Truth Response}: The ideal or reference response expected.\\[1ex]
    \quad \textbf{Your task is to judge the relevance of the Predicted Response in addressing the User Query, considering the Context Conversation.} Relevance measures how well the Predicted Response aligns with the user's needs and intentions as expressed in the User Query and Context Conversation.\\[1ex]
    \quad \textbf{Instructions}\\
    \quad 1. \textbf{Analyze the Context Conversation} to understand the background and the user's intent.\\
    \quad 2. \textbf{Evaluate the Predicted Response} to determine how effectively it addresses the User Query in the given context.\\
    \quad 3. \textbf{Compare the Predicted Response with the Ground Truth Response} to gauge its alignment with the expected answer.\\[1ex]
    \quad \textbf{Rating Scale (1 to 5)}\\
    \quad - \textbf{5}: The Predicted Response is highly relevant, directly addressing the User Query with precise and contextually appropriate information.\\
    \quad - \textbf{4}: The Predicted Response is mostly relevant but may lack some minor details or nuances present in the Ground Truth Response.\\
    \quad - \textbf{3}: The Predicted Response is moderately relevant, addressing the main aspects of the User Query but missing significant details or context.\\
    \quad - \textbf{2}: The Predicted Response is minimally relevant, only partially addressing the User Query and lacking substantial alignment with the context.\\
    \quad - \textbf{1}: The Predicted Response is irrelevant, failing to address the User Query and not aligned with the context.\\[1ex]
    \quad \textbf{Output Requirements}\\
    \quad For each evaluation, provide:\\
    \quad - A \textbf{numerical rating} (1 to 5) based on the criteria above.\\
    \quad - A \textbf{brief explanation} (1 sentence) justifying your rating, highlighting specific strengths or weaknesses in the Predicted Response.\\[1ex]
    \quad \textbf{Components}\\
    \quad - \textbf{Context Conversation}: \{context\_conversation\}\\
    \quad - \textbf{User Query}: \{data\_point['query']\}\\
    \quad - \textbf{Ground Truth Response}: \{data\_point['gt\_response']\}\\
    \quad - \textbf{Predicted Response}: \{data\_point['prediction']\}\\
    \quad - \textbf{Predicted Rating:}

\end{monopara}
\end{tcolorbox}
\caption{Relevance Evaluation Prompt}
\label{fig:relevence_jury_prompt}
\end{figure*}


\begin{figure*}[!ht]
\noindent
\begin{tcolorbox}[
  colback=white,
  colframe=black,
  arc=2mm,           
  boxrule=0.5pt,     
  width=\textwidth,  
  toptitle=1pt,
  bottomtitle=1pt,
  title=\bfseries Sufficiency Prompt used for Judge Models.]
  \begin{monopara}
  
    You are a spiritual counselor.\\[1ex]
    \quad You will be provided with the following components:\\
    \quad 1. \textbf{Context Conversation}: The preceding dialogue or background information leading up to the user's query.\\
    \quad 2. \textbf{User Query}: The specific question or statement from the user.\\
    \quad 3. \textbf{Predicted Response}: The response generated by the model.\\
    \quad 4. \textbf{Ground Truth Response}: The ideal or reference response expected.\\
    \quad 5. \textbf{Key Teachings List}: A set of key spiritual concepts, values, or teachings relevant to the context.\\[1ex]
    \quad Your task is to judge the Sufficiency of the Predicted Response in addressing the User Query, considering the Context Conversation and its alignment with the Key Teachings List. Sufficiency measures whether the response provides enough detail and depth to comprehensively address the user's query and the context while reflecting key spiritual principles.\\[1ex]
    \quad \textbf{Instructions}\\
    \quad 1. Analyze the Context Conversation to understand the user's intent and concerns.\\
    \quad 2. Evaluate the Predicted Response to determine if it provides sufficient detail, depth and comprehensiveness to address the User Query in the given context.\\
    \quad 3. Compare the Predicted Response with the Ground Truth Response to assess alignment and completeness in addressing the User Query.\\
    \quad 4. Refer to the Key Teachings List to ensure the Predicted Response incorporates relevant spiritual teachings and values to enhance depth.\\[1ex]
    \quad \textbf{Rating Criteria}\\
    \quad Use the following rating scale (1 to 5) to assess sufficiency:\\
    \quad - \textbf{5 (Highly Sufficient)}: The Predicted Response comprehensively addresses the User Query with detailed, in-depth information and effectively integrates key spiritual teachings.\\
    \quad - \textbf{4 (Mostly Sufficient)}: The Predicted Response addresses the User Query well but may lack minor details or nuances that could enhance comprehensiveness.\\
    \quad - \textbf{3 (Moderately Sufficient)}: The Predicted Response provides some useful information but misses significant details or depth needed for a comprehensive response.\\
    \quad - \textbf{2 (Minimally Sufficient)}: The Predicted Response addresses the User Query partially, with noticeable gaps in detail, depth, or relevance to key spiritual teachings.\\
    \quad - \textbf{1 (Not Sufficient)}: The Predicted Response fails to provide enough detail or depth to address the User Query and does not incorporate key spiritual teachings.\\[1ex]
    \quad \textbf{Output Requirements}\\
    \quad For each evaluation, provide:\\
    \quad - A numerical rating (1 to 5) based on the criteria above.\\
    \quad - A brief explanation (1 sentence) justifying your rating, including specific references to the Key Teachings List and how they are (or are not) reflected in the Predicted Response.\\[1ex]
    \quad \textbf{Components}\\
    \quad - \textbf{Context Conversation}: \{context\_conversation\}\\
    \quad - \textbf{User Query}: \{data\_point['query']\}\\
    \quad - \textbf{Ground Truth Response}: \{data\_point['gt\_response']\}\\
    \quad - \textbf{Predicted Response}: \{data\_point['prediction']\}\\
    \quad - \textbf{Key Teaching List}: \{key\_teachings\}\\
    \quad - \textbf{Predicted Rating:}
  \end{monopara}
\end{tcolorbox}
\caption{Sufficiency Evaluation Prompt}
\label{fig:sufficiency_prompt}
\end{figure*}


\end{document}